\documentclass[10pt,twocolumn,letterpaper]{article}
\PassOptionsToPackage{table,x11names}{xcolor}
\usepackage[pagenumbers]{cvpr}
\usepackage[utf8]{inputenc}
\usepackage[T1]{fontenc}
\usepackage{amsfonts}
\usepackage{nicefrac}
\usepackage{microtype}
\usepackage{pifont}
\usepackage{enumitem}
\usepackage{multirow}
\usepackage{stfloats}
\newcommand{\blockcomment}[1]{}
\definecolor{cvprblue}{rgb}{0.21,0.49,0.74}
\usepackage[pagebackref,breaklinks,colorlinks,allcolors=cvprblue]{hyperref}
\newcommand{\cmt}[1]{}
\providecommand{\comment}[1]{}
\renewcommand{\comment}[1]{}
\def\confName{CVPR}
\def\confYear{2027}
\title{\vspace{-0.8em}SCALER: Breaking the Latent Scaling Bottleneck via Single-Pass Context-Anchored Latent Visual Reasoning\vspace{-0.5em}}
\author{%
Chenfeng Wang\textsuperscript{1,2}\quad
Wei Zhai\textsuperscript{1,$\dagger$}\quad
Wei He\textsuperscript{2}\quad
Xuhan Zhu\textsuperscript{2}\quad
Chunpeng Zhou\textsuperscript{2}\\[-0.1em]
Song Yan\textsuperscript{1}\quad
Qizhen Li\textsuperscript{2}\quad
Zeyu Zhang\textsuperscript{1,2}\quad
Shuai Guo\textsuperscript{1,2}\quad
Yufei Zheng\textsuperscript{1,2}\\[-0.1em]
Chengjun Yu\textsuperscript{1,2}\quad
Fan Lu\textsuperscript{1}\quad
Yang Cao\textsuperscript{1}\quad
Pengfei Yu\textsuperscript{2,$\dagger$}\quad
Zheng-Jun Zha\textsuperscript{1}\\[0.2em]
\textsuperscript{1}University of Science and Technology of China\qquad
\textsuperscript{2}Li Auto Inc.
}

\begin{document}
\twocolumn[{%
\renewcommand\twocolumn[1][]{#1}%
\maketitle
\vspace{-3.6em}
\begin{center}
  \centering
  \captionsetup{type=figure}
  \includegraphics[width=\linewidth]{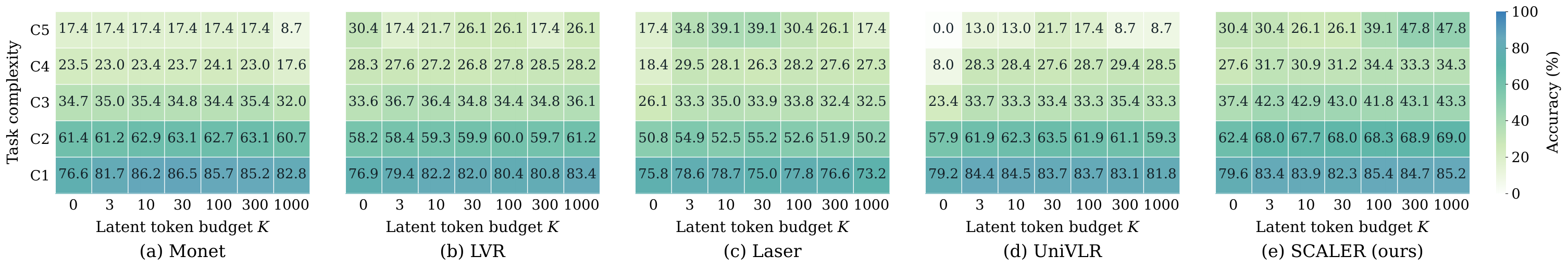}
  \captionof{figure}{\textbf{The Latent Scaling Bottleneck Across Task Complexity (C1--C5) and Latent Token Budgets ($K \in [0, 1000]$).}
    Accuracy (\%) across five pooled benchmarks stratified into five complexity levels. Existing methods ((a) Monet, (b) LVR, (c) LASER, and (d) UniVLR) only exhibit positive length scaling in the shallow lower-left regime ($K \leq 30$, low complexity C1--C2), but stagnate or degrade in the upper-right regime of higher task complexity (C3--C5) and longer sequences ($K \geq 100$). In contrast, (e) \textbf{SCALER} sustains positive scaling across long latent budgets and achieves substantial gains on high-complexity tasks (C4--C5).}
  \label{fig:confusion_matrix}
\end{center}%
\vspace{0.0em}
}]
\begin{abstract}
Longer chains of thought substantially improve language reasoning on complex problems, inspiring efforts to scale visual reasoning via continuous latents. However, evaluating visual latent reasoning across complexity and sequence length reveals a fundamental \textbf{Latent Scaling Bottleneck}: existing methods yield modest gains only within short budgets ($\leq$30 tokens) on simple tasks, yet stagnate or \emph{degrade} on harder tasks and longer chains. Controlled diagnostics trace this bottleneck to two root causes. First, \textbf{Latent Representation Saturation} under autoregressive conditioning causes step-wise cosine novelty to decay exponentially and saturate after $\sim$30 steps. Second, \textbf{Over-Compressed Supervision Targets} from heavily pooled ($\geq$128$\times$) image embeddings lose fine-grained spatial structure and act like meaningless padding. We propose \textbf{SCALER} (\textbf{S}ingle-Pass \textbf{C}ontext-\textbf{A}nchored \textbf{L}at\textbf{e}nt Visual \textbf{R}easoning), whose lightweight \emph{detransformer} synthesizes spatially aligned auxiliary tokens from full-sequence hidden states \emph{in a single forward pass}, anchoring each token to the original visual space. Coupled with three-stage SFT and \textbf{ALPO} (\textbf{A}uxiliary \textbf{L}atent \textbf{P}olicy \textbf{O}ptimization) for differentiable RL, SCALER extends effective latent length by over 30$\times$, achieves state-of-the-art open-source accuracy (e.g., \textbf{57.84\%} on MME-RealWorld-Lite, +12.09\% over backbone, +10.47\% over GPT-6-luna), with strong OOD generalization. Our project page: {\small\url{https://bonjour-codedev.github.io/SCALER/}}.
\end{abstract}

\section{Introduction}

The ``compute-for-performance'' scaling paradigm has transformed language reasoning: longer chain-of-thought (\textbf{CoT}) sequences substantially unlock stronger capabilities on complex multi-step problems~\cite{chen2025towards,jaech2024openai,guo2025deepseek,yeo2025demystifying}. This naturally raises a fundamental question: \emph{can visual reasoning similarly scale with longer latent reasoning chains to solve harder visual problems?} Recent studies~\cite{wang2025monetreasoninglatentvisual,li2025latent,yang2025mirage,dong2025ilvr,cheng2026hybrid,wang2026laser,jiang2026univlr} in vision-language models (\textbf{VLMs})~\cite{bai2025qwen2,cheng2025visual,chen2025mindgpt,jian2025look,mavors,comanici2025gemini,wang2025internvl3} explore this direction by generating continuous visual latent tokens as intermediate reasoning steps, expecting longer latent sequences to accumulate richer visual evidence.

Despite promising gains on standard benchmarks, these studies overlook a critical question: \emph{does visual latent reasoning genuinely scale as both sequence length and task complexity increase?} Evaluating existing methods across a two-dimensional grid of \emph{task complexity} (C1--C5) and \emph{latent token budget} ($K \in \{0, 3, 10, 30, 100, 300, 1000\}$) in Figure~\ref{fig:confusion_matrix}, we uncover a fundamental \textbf{Latent Scaling Bottleneck}. Existing methods~\cite{li2025latent,wang2025monetreasoninglatentvisual,dong2025ilvr,cheng2026hybrid,wang2026laser} indeed exhibit positive length scaling in a narrow \textbf{shallow regime} of small budgets ($K \leq 30$) and low-complexity tasks (C1--C2), which explains the gains previously reported under short budgets ($K \leq 16$). Yet as task complexity rises to C3--C5 and sequences extend beyond 30 tokens, precisely where deeper reasoning is needed, existing methods hit a sharp ceiling: accuracy either stagnates (LVR) or systematically degrades as more latent tokens are generated (Monet, LASER, UniVLR).

\begin{figure*}[!t]
  \centering
  \includegraphics[width=\linewidth]{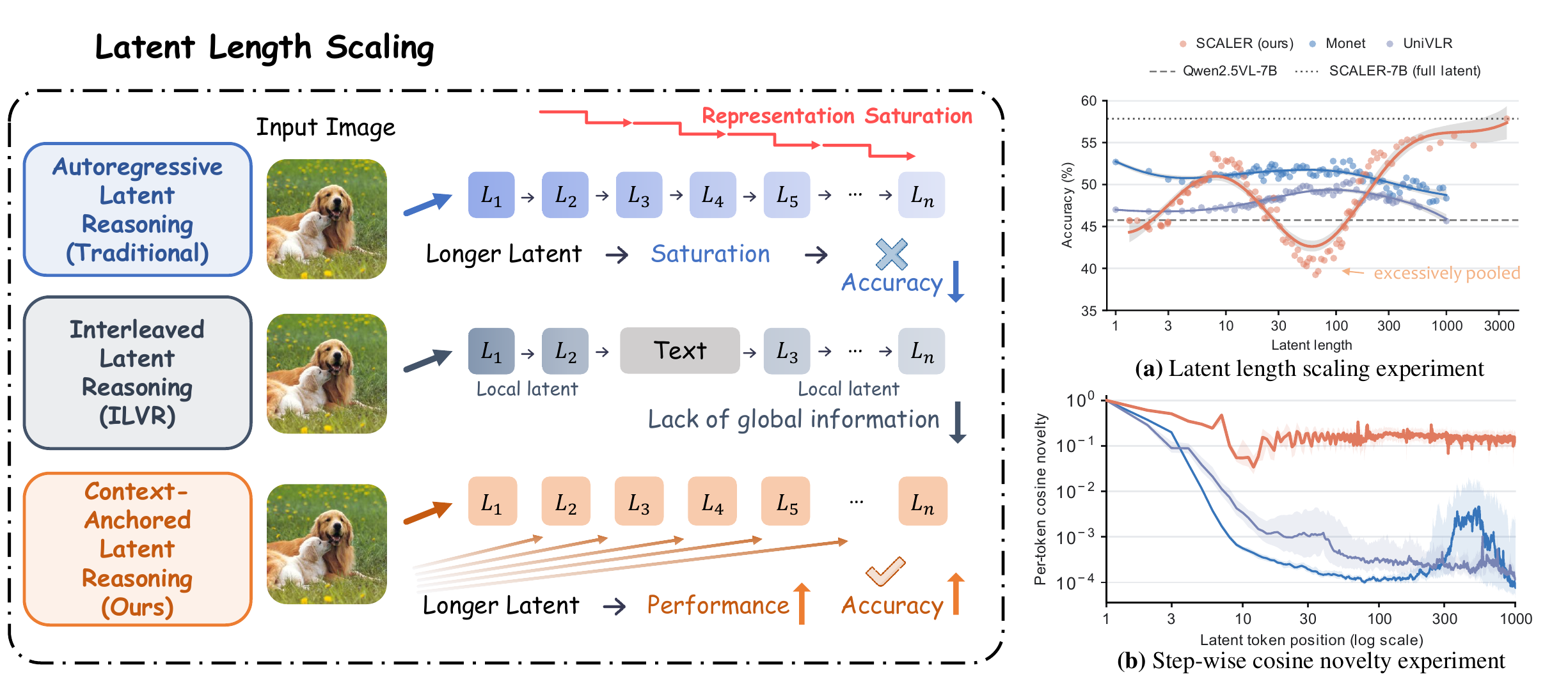}
  \caption{\textbf{Root Cause Diagnosis and Our Single-Pass Context-Anchored Solution.}
    \textbf{Left:} Paradigm comparison. In conventional autoregressive latent reasoning, latent variable $L_n$ depends recursively on $L_{n-1}$, causing visual semantics to decay and representations to saturate (\textbf{Accuracy $\downarrow$}). Interleaved latent reasoning (ILVR) inserts short local latent spans between text tokens but lacks global high-resolution visual grounding. In \textbf{SCALER}, each auxiliary token independently anchors to the original visual semantic space in a single forward pass, preventing information decay (\textbf{Accuracy $\uparrow$}).
    \textbf{Right (a):} Latent length scaling on MME-RealWorld-Lite (scatter points, $\log_{10}$-length local quadratic fits, and 95\% bootstrap bands): SCALER recovers from moderate-pooling dips to peak at full length, whereas Monet and UniVLR plateau and degrade.
    \textbf{Right (b):} Step-wise cosine novelty on MME-RealWorld-Lite across 1K tokens: SCALER sustains high novelty ($10^{-1}\text{--}10^{0}$) throughout the sequence, whereas Monet and UniVLR decay exponentially to $10^{-3}\text{--}10^{-4}$ after $t \approx 30$.}
  \label{fig:motivation}
\end{figure*}

\textbf{Why do existing methods hit this Latent Scaling Bottleneck?} We dissect the underlying failure mechanism through three complementary diagnostics: \emph{(1)~benchmark length scaling} (Figure~\ref{fig:motivation}(a)), which shows that longer chains systematically hurt or stall existing methods; \emph{(2)~meaningless padding interventions} (Figure~\ref{fig:filling}), which reveal that existing decoders either bypass latent semantics altogether (UniVLR, LVR, LASER) or suffer from destructive structural interference (Monet); and \emph{(3)~step-wise cosine novelty analysis} (Figure~\ref{fig:motivation}(b)), which tracks token redundancy across 1000 steps. Together, these diagnostics isolate two root causes. First, \textbf{Latent Representation Saturation}: because autoregressive prediction recursively conditions each token on its own prior outputs, original visual information rapidly erodes along the chain and step-wise novelty saturates after roughly 30 steps, while interleaved variants~\cite{dong2025ilvr} lose global scene context. Consequently, later tokens merely repeat earlier states and cannot supply new visual evidence for high-complexity tasks (C3--C5). Second, \textbf{Over-Compressed Supervision Targets}: to fit short chains, prior methods heavily pool ($\geq$128$\times$) auxiliary images into a few target tokens, erasing fine-grained spatial structure and reducing targets to noisy distractors.

\begin{figure*}[!t]
  \centering
  \includegraphics[width=\textwidth]{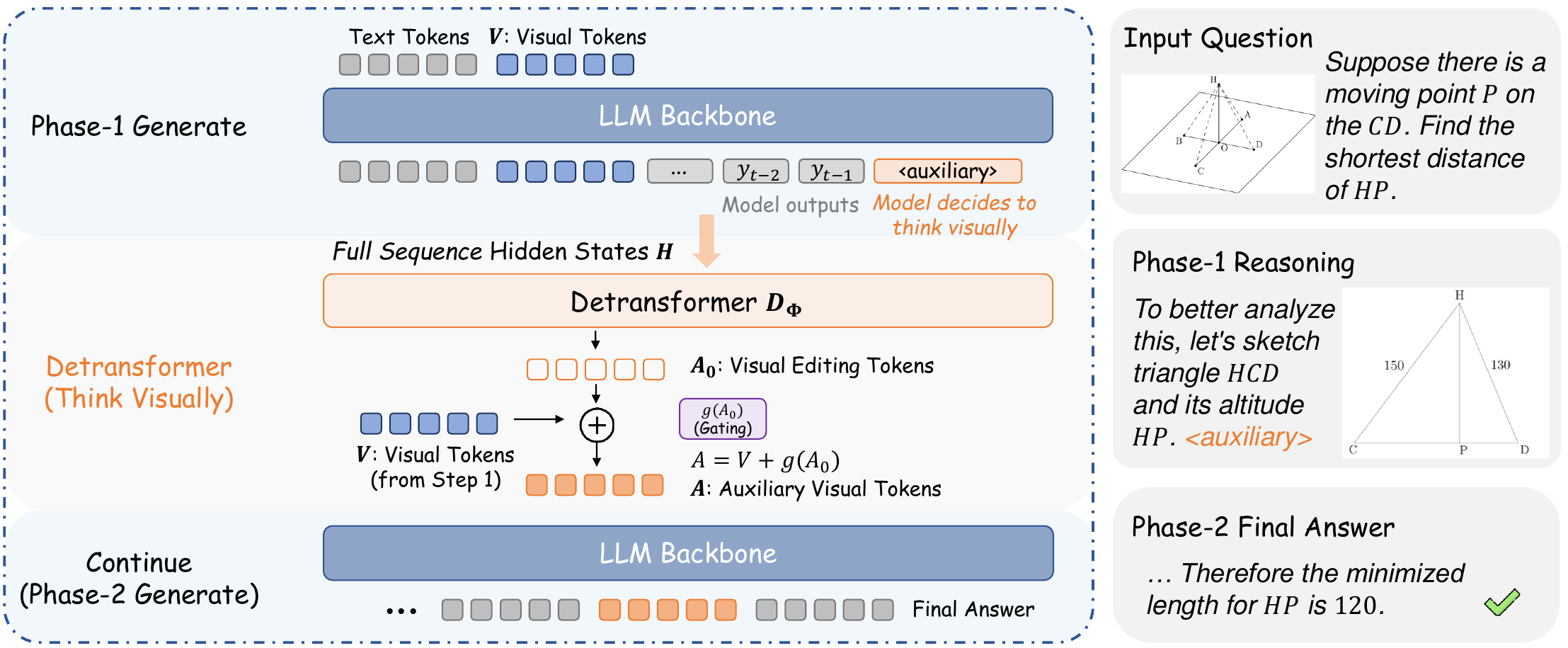}
  \caption{\textbf{Overview of the SCALER Inference Pipeline.} \textbf{Left:} Three-stage inference flow: (1)~\emph{Phase-1 Generate}: the LLM backbone processes text tokens and visual tokens $V$ until autonomously emitting the \texttt{<auxiliary>} trigger; (2)~\emph{Detransformer (Think Visually)}: full-sequence hidden states $H$ are mapped by $D_\phi$ in a single forward pass to visual editing tokens $A_0$, which are residually fused with $V$ via gating $g(A_0)$ to form auxiliary visual tokens $A = V + g(A_0)$; (3)~\emph{Continue (Phase-2 Generate)}: the LLM resumes decoding with $A$ in context to produce the final answer. \textbf{Right:} An illustrative geometry reasoning example where SCALER triggers \texttt{<auxiliary>} to synthesize the auxiliary diagram of $\triangle HCD$ and its altitude $HP$.}
  \label{fig:method}
\end{figure*}

These findings yield a clear design principle: breaking the Latent Scaling Bottleneck requires each auxiliary token to \emph{independently anchor to the original visual semantic space} rather than relying on recursive step-by-step transmission. We therefore propose \textbf{SCALER} (\textbf{S}ingle-Pass \textbf{C}ontext-\textbf{A}nchored \textbf{L}at\textbf{e}nt Visual \textbf{R}easoning). Built on the standard ViT--Projector--LLM architecture~\cite{liu2023visual,li2024llava}, SCALER introduces a lightweight \emph{detransformer} branch that maps full-sequence LLM hidden states \emph{in a single forward pass} to a spatially aligned auxiliary visual sequence, residually fusing it with original visual features to preserve fine-grained spatial structure (lower-left of Figure~\ref{fig:motivation}). Being \textbf{context-anchored} ensures that auxiliary tokens remain grounded throughout generation and downstream reasoning: spatially aligned visual features provide dense supervision targets so every token encodes distinct spatial semantics; a three-stage SFT curriculum with teacher-forcing annealing bridges ground-truth supervision and self-generated latents; and \textbf{ALPO} (\textbf{A}uxiliary \textbf{L}atent \textbf{P}olicy \textbf{O}ptimization) recomputes continuous auxiliary latents with preserved gradients during RL updates, allowing outcome-driven policy gradients to backpropagate through the Phase-2 reasoning context into the detransformer and prefix LLM layers.

By eliminating representation saturation, SCALER sustains high step-wise cosine novelty across 1000 tokens (Figure~\ref{fig:motivation}(b)), recovers from moderate-pooling dips to peak at full spatial resolution (Figure~\ref{fig:motivation}(a)), and unlocks consistent length scaling on high-complexity C4--C5 tasks (Figure~\ref{fig:confusion_matrix}(e)), extending effective latent sequence length by over 30$\times$ across visual domains. Our main contributions are:
\begin{itemize}
\item We uncover the \emph{Latent Scaling Bottleneck} across task complexity and sequence length, trace its root causes to \emph{Latent Representation Saturation} (quantified via step-wise cosine novelty) and over-compressed supervision, and propose \textbf{SCALER}, a single-pass paradigm whose lightweight detransformer synthesizes spatially aligned auxiliary tokens anchored to the original visual space.
\item We design a progressive three-stage SFT curriculum with teacher-forcing annealing alongside \textbf{ALPO}, a reinforcement learning algorithm with differentiable latent recomputation that jointly optimizes latent triggering, continuous representation synthesis, and multi-step reasoning.
\item SCALER extends effective latent sequence length by over 30$\times$, achieving state-of-the-art open-source accuracy on V*\,Bench (\textbf{84.29\%}, +9.42\% over GPT-6-luna~\cite{openai2026gpt6}), HRBench4K/8K~\cite{wang2025divide_hrbench}, MME-RealWorld-Lite~\cite{zhang2024mme}, and OOD VisualPuzzles~\cite{song2025visualpuzzles}.
\end{itemize}

\section{Related Work}
\textbf{Think with Images.} Multimodal reasoning methods enrich visual understanding along two main directions: eliciting longer textual CoT over static images via SFT or RL~\cite{xu2025llava,peng2025lmm,wei2025advancing,jiang2025vlm,yu2025perception,wang2025visualprm}, or explicitly injecting visual operations into the reasoning loop, such as region grounding and cropping~\cite{wang2025pixel,zhang2025adaptive,chen2026streamingclaw,huang2025visualtoolagent}, invoking external visual tools to draw or edit~\cite{qiao2025vthinker,chen2025learning,fu2025refocus,hu2024visual,qi2024cogcom,su2025openthinkimg,zhang2025thyme,zhao2025pyvision,zhou2025reinforced,zheng2025deepeyes,hong2026deepeyesv2}, and synthesizing intermediate visual states~\cite{he2025diffthinkergenerativemultimodalreasoning,chern2025thinking,li2025imagine,xu2025visualplanningletsthink}. Despite strong localization ability, these pipelines remain constrained by symbolic text bottlenecks or rely on multi-turn external tool execution with high deployment complexity and latency.

\textbf{Latent Space Reasoning in VLMs.} Shifting intermediate reasoning from discrete tokens to continuous hidden states has gained traction in both NLP~\cite{coconut,butt2025softtokenshard,geiping2025scaling,wang2025synadapt,wei2025sim} and VLMs. Monet~\cite{wang2025monetreasoninglatentvisual}, LVR~\cite{li2025latent}, and Mirage~\cite{yang2025mirage} align autoregressively generated latent vectors with auxiliary or cropped image embeddings, whereas SkiLa~\cite{tong2025sketchinlatentselicitingunifiedreasoning} interleaves textual reasoning with latent visual sketches supervised by visual-feature reconstruction. More recently, ILVR~\cite{dong2025ilvr} and HyLaR~\cite{cheng2026hybrid} interleave short latent spans with text reasoning (sacrificing global scene context; Appendix~\ref{app:ilvr_comparison}), LASER~\cite{wang2026laser} applies dynamic-window latent superposition, UniVLR~\cite{jiang2026univlr} and LUT~\cite{kang2026lut} couple latent decoding with RL optimization, and ProLaViT~\cite{li2026prolavit} and CoLVR~\cite{ding2026colvr} introduce diversity or contrastive regularizers. Although recent studies~\cite{viveiros2026holding,li2026imagination,ding2026colvr,wang2026laser,li2026prolavit} note that visual latent tokens are prone to representation collapse, they largely apply local regularizers within short chains without addressing the root causes: \emph{recursive autoregressive conditioning} and \emph{over-compressed ($\geq$128$\times$ pooled) supervision targets}. Moreover, existing RL-based latent methods typically treat generated latents as detached fixed context during policy updates, whereas SCALER couples single-pass context-anchored synthesis with differentiable latent recomputation in ALPO to sustain high per-token novelty across 1000+ latent tokens.

\begin{figure*}[!t]
  \centering
  \includegraphics[width=\textwidth]{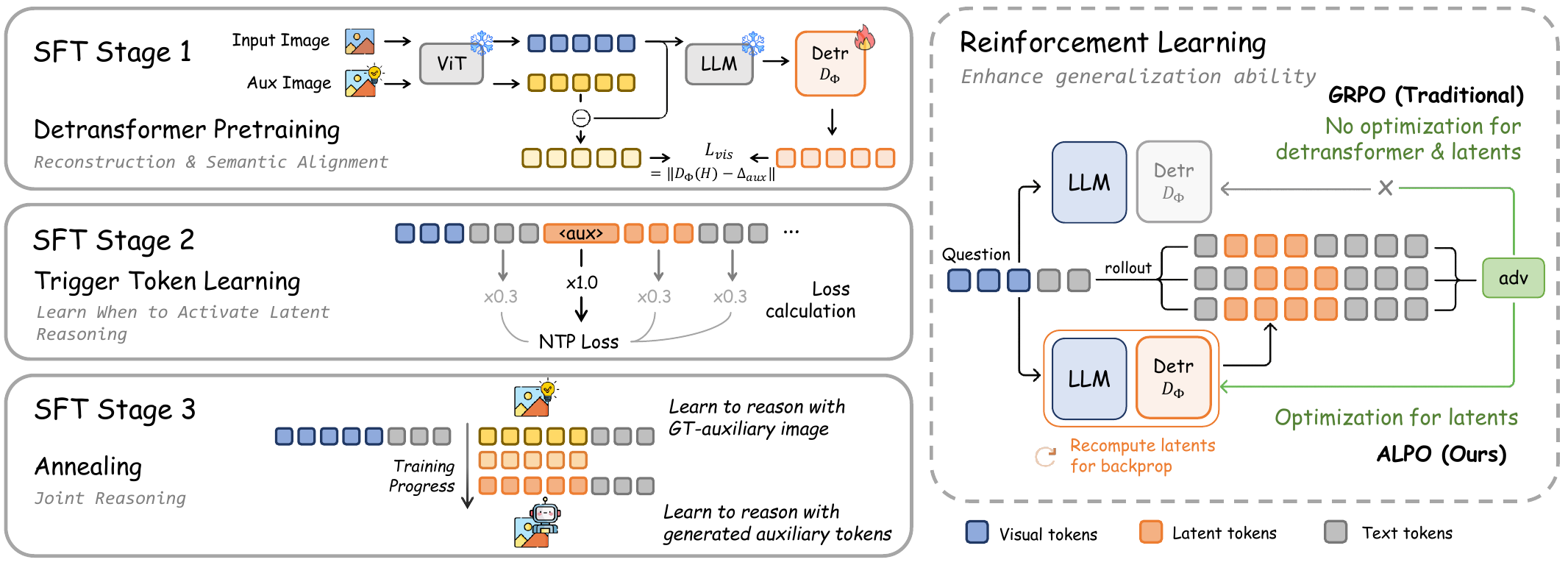}
  \caption{\textbf{Overview of the SCALER Training Pipeline.}
    \textbf{Stage 1:} Detransformer pretraining via delta-feature reconstruction.
    \textbf{Stage 2:} Trigger learning via weighted NTP.
    \textbf{Stage 3:} Joint reasoning with teacher-forcing annealing.
    \textbf{RL:} ALPO with differentiable latent recomputation and outcome rewards.}
  \label{fig:training}
\end{figure*}

\section{Method}
\subsection{Overview}
\label{sec:inference}
Built on the standard ViT--Projector--LLM architecture, SCALER introduces a lightweight \emph{detransformer} branch $D_\phi$ that synthesizes \textbf{auxiliary visual tokens} aligned with the input spatial resolution from the LLM's \emph{full-sequence hidden states} in a single forward pass, fusing them with original visual features to serve subsequent language reasoning. This \textbf{single-pass, context-anchored} latent generation is the key to scaling latent length without performance degradation. Specifically, as illustrated in Figure~\ref{fig:method}, inference proceeds in three stages:
\textbf{(1)~Phase-1 Generate:} Given the input question tokens and visual tokens $V$ (encoded from the input image via the ViT--Projector), the LLM backbone generates Phase-1 reasoning tokens $y_{1:t}$ until its policy autonomously emits the trigger token \texttt{<auxiliary>} to initiate visual thinking (whether to trigger is determined by the model's policy and may occur zero or multiple times; e.g., sketching $\triangle HCD$ and altitude $HP$ in Figure~\ref{fig:method}, right);
\textbf{(2)~Detransformer (Think Visually):} The full-sequence hidden states $H$ at the trigger step are extracted and fed into the detransformer $D_\phi$ to synthesize visual editing tokens $A_0$ in a single forward pass, which are residually fused with the original visual tokens $V$ via gating $g(A_0)$ to form auxiliary visual tokens $A = V + g(A_0)$;
\textbf{(3)~Continue (Phase-2 Generate):} The LLM resumes Phase-2 decoding with $A$ in context to produce the final answer.

Formally, let $V = E_{\mathrm{vis}}(I) \in \mathbb{R}^{N_v \times d_v}$ and $X_v = P(V) \in \mathbb{R}^{N_v \times d}$ denote the visual feature sequence and projected visual token sequence for input image $I$, where $N_v$ is the number of visual tokens and $d$ is the LLM hidden dimension. Upon emitting \texttt{<auxiliary>} after prefix $y_{1:t}$, full-sequence hidden states $H$ from layer $\ell$ are mapped by $D_\phi$ to editing tokens $A_0$:
\begin{equation}
  \begin{aligned}
  H &= f_\theta^{(\ell)}([X_v; q; y_{1:t}; \texttt{<auxiliary>}]) \in \mathbb{R}^{L \times d}, \\
  A_0 &= D_\phi(H) \in \mathbb{R}^{N_a \times d_v},
  \end{aligned}
\end{equation}
where $q$ is the question, $L$ is the full sequence length, and $N_a = N_v$ matches the input visual token count.

\textbf{Detransformer Architecture.}
Implemented as a lightweight two-layer Transformer~\cite{vaswani2017attention} operating on $H$, $D_\phi$ applies layer normalization, multi-head self-attention, and feed-forward blocks (Appendix~\ref{app:detransformer}), indexing the $N_v$ visual positions to output $A_0$ at the input spatial resolution. The output is fused with original visual features $V$ via gated residual addition:
\begin{equation}
  A = V + g(A_0),
  \qquad
  g(A_0) = \alpha\,A_0,\quad \alpha \in [0,1],
\end{equation}
where $\alpha$ controls editing strength. The fused features $A$ are projected into the LLM context to resume Phase-2 decoding:
\begin{equation}
  y_{t+1}=\text{decode}\big(f_\theta([X_v; q; y_{1:t}; P(A)])\big),
  \label{eq:inject}
\end{equation}
where $\text{decode}(\cdot)$ projects hidden states through the language model head to sample the next token.

\subsection{Supervised Fine-Tuning (SFT)}
\label{sec:sft_data}
The training set consists of geometrically aligned tuples $(I, x, y^{(1)}, I_{\text{aux}}, y^{(2)})$ constructed via SIFT~\cite{lowe1999object,lowe2004distinctive} alignment, CoT distillation, and necessity filtering (Appendix~\ref{app:sft_data}). Direct end-to-end training suffers from \emph{implicit distribution mismatch}: ground-truth auxiliary visuals are used during training, whereas inference relies on \emph{self-generated} latents. We therefore design a progressive \textbf{three-stage SFT} curriculum (Figure~\ref{fig:training}): (i)~detransformer pretraining for residual reconstruction, (ii)~weighted NTP for trigger acquisition, and (iii)~joint reasoning with teacher-forcing annealing.

\begin{table*}[!t]
    \centering
    \caption{\textbf{Performance on real-world perception and reasoning benchmarks.} The best-performing open-source model for each dataset is highlighted in \textbf{bold}, while the second-best is marked with an \underline{underline}.}
    \label{tab:exp-main}
    \renewcommand{\arraystretch}{1.10}
    \setlength{\tabcolsep}{5pt}
    \resizebox{\textwidth}{!}{%
    \begin{tabular}{lcccccccccccc}
        \toprule
        \multirow{2}{*}{\textbf{Methods}} & \multicolumn{3}{c}{\textbf{V$^*$}} & \multicolumn{3}{c}{\textbf{HRBench-4K}} & \multicolumn{3}{c}{\textbf{HRBench-8K}} & \multicolumn{3}{c}{\textbf{MME-RealWorld-Lite}} \\
        \cmidrule(lr){2-4}\cmidrule(lr){5-7}\cmidrule(lr){8-10}\cmidrule(lr){11-13}
        & Overall & Attribute & Spatial & Overall & FSP & FCP & Overall & FSP & FCP & Overall & Reasoning & Perception \\
        \midrule
        \rowcolor[HTML]{EBEBFF}
        \multicolumn{13}{c}{\textit{Textual Reasoning}} \\
        GPT-4o$_{\text{(2024)}}$~\cite{gpt_4o} & 67.5* & 72.2* & 60.5* & 59.0* & 70.0* & 48.0* & 55.5* & 62.0* & 49.0* & 52.0* & 48.3* & 54.4* \\
        GPT-6-luna$_{\text{(2026)}}$~\cite{openai2026gpt6} & 74.87 & 77.39 & 71.05 & 72.88 & 83.25 & 62.50 & 62.25 & 65.00 & 59.50 & 47.37 & 44.80 & 49.02 \\
        Qwen2.5-VL-7B$_{\text{(2025)}}$~\cite{bai2025qwen2} & 78.53 & 80.00 & 76.31 & 71.50 & 83.00 & 60.00 & 63.75 & 73.75 & 53.75 & 45.75 & 39.73 & 49.62 \\
        InternVL3-8B$_{\text{(2025)}}$~\cite{zhu2025internvl3} & 73.82 & 70.43 & 78.95 & 70.25 & 77.25 & \textbf{63.25} & 66.75 & 77.75 & 55.75 & 48.72 & 44.53 & 51.41 \\
        \midrule
        \rowcolor[HTML]{F9E6EC}
        \multicolumn{13}{c}{\textit{Tool-based Visual Reasoning}} \\
        PixelReasoner$_{\text{(2025)}}$~\cite{wang2025pixel} & 80.6* & 83.5* & 76.3* & 72.9* & 86.0* & 60.3* & 66.9* & 80.0* & 54.3* & 49.7* & 44.5* & 53.1* \\
        DeepEyes$_{\text{(2026)}}$~\cite{zheng2025deepeyes} & 83.25 & 84.35 & 81.58 & 71.25 & 83.75 & 58.75 & 65.13 & 77.00 & 53.25 & 54.28 & \underline{50.53} & 56.63 \\
        \midrule
        \rowcolor[HTML]{E6FFE6}
        \multicolumn{13}{c}{\textit{Visual Latent Reasoning}} \\
        LVR$_{\text{(2025)}}$~\citep{li2025latent} & 81.15 & 84.35 & 76.32 & 69.50 & 82.25 & 56.75 & 62.25 & 73.75 & 50.75 & 49.66 & 44.27 & 53.12 \\
        Monet$_{\text{(2026)}}$~\citep{wang2025monetreasoninglatentvisual} & \underline{83.77} & \underline{85.22} & 81.58 & 73.13 & 86.75 & 59.50 & 67.13 & \underline{80.25} & 54.00 & 52.63 & 46.93 & 56.29 \\
        HyLaR$_{\text{(2026)}}$~\citep{cheng2026hybrid} & 80.10 & 83.48 & 75.00 & 73.25 & \textbf{93.25} & 53.25 & \textbf{69.38} & \textbf{87.50} & 51.25 & 52.74 & 47.07 & 56.37 \\
        LASER$_{\text{(2026)}}$~\citep{wang2026laser} & 76.96 & 80.00 & 72.37 & 73.25 & 88.00 & 58.50 & 66.13 & 78.75 & 53.50 & 50.96 & 45.73 & 54.32 \\
        UniVLR$_{\text{(2026)}}$~\citep{jiang2026univlr} & 82.7* & 83.5* & 81.6* & 73.3* & 86.0* & 60.5* & 68.8* & 78.8* & \underline{58.8*} & 50.7* & 44.7* & 54.5* \\
        LUT-7B$_{\text{(2026)}}$~\citep{kang2026lut} & \textbf{84.29*} & \textbf{86.09*} & 81.58* & 72.62* & 86.50* & 58.75* & 68.12* & 80.00* & 56.25* & \underline{54.82*} & 47.33* & \textbf{59.62*} \\
        \textbf{SCALER-SFT (Ours)} & 82.72 & 82.61 & \underline{82.89} & \underline{74.25} & 87.00 & 61.50 & 66.75 & 75.75 & 57.75 & 52.84 & 50.27 & 54.49 \\
        \textbf{SCALER-7B (Ours)} & \textbf{84.29} & 84.35 & \textbf{84.21} & \textbf{75.25} & \underline{88.75} & \underline{61.75} & \underline{68.88} & 76.50 & \textbf{61.25} & \textbf{57.84} & \textbf{55.73} & \underline{59.20} \\
        \rowcolor[HTML]{ECECEC}
        \textit{$\Delta$} (vs Qwen2.5-VL-7B) & \textcolor[rgb]{0,0.6,0}{+5.76} & \textcolor[rgb]{0,0.6,0}{+4.35} & \textcolor[rgb]{0,0.6,0}{+7.90} & \textcolor[rgb]{0,0.6,0}{+3.75} & \textcolor[rgb]{0,0.6,0}{+5.75} & \textcolor[rgb]{0,0.6,0}{+1.75} & \textcolor[rgb]{0,0.6,0}{+5.13} & \textcolor[rgb]{0,0.6,0}{+2.75} & \textcolor[rgb]{0,0.6,0}{+7.50} & \textcolor[rgb]{0,0.6,0}{+12.09} & \textcolor[rgb]{0,0.6,0}{+16.00} & \textcolor[rgb]{0,0.6,0}{+9.58} \\
        \bottomrule
    \end{tabular}%
    }
\end{table*}

\textbf{Stage 1: Detransformer Pretraining.}
Only $D_\phi$ is trained while $f_\theta$ remains frozen, preventing an unaligned visual branch from perturbing the LLM's semantic representations. Rather than reconstructing the entire image from scratch, $D_\phi$ predicts the \textbf{visual editing residual} $\Delta_{\text{aux}}$:
\begin{equation}
  \begin{aligned}
  \Delta_{\text{aux}} &= E_{\mathrm{vis}}(I_{\text{aux}}) - E_{\mathrm{vis}}(I), \\
  \mathcal{L}_{\text{vis}} &= \| D_\phi(H) - \Delta_{\text{aux}} \|_2^2.
  \end{aligned}
\end{equation}
During training, $H$ is extracted from the full prefix $(I, x, y^{(1)})$ preceding \texttt{<auxiliary>}, ensuring that $D_\phi$ is modulated by the intermediate textual reasoning intent rather than performing unconstrained visual editing.

\textbf{Stage 2: Learning the Trigger Token.}
We next train the LLM to autonomously invoke visual thinking at appropriate reasoning steps via \textbf{weighted next-token prediction} (NTP):
\begin{equation}
\begin{aligned}
\mathcal{L}_{\text{ntp}} &= -\sum_{t} w_t \log \pi_\theta(y_t \mid y_{<t}, V, x_{1:T}), \\
w_t &= \begin{cases}
w_{\text{aux}}, & y_t = \text{\texttt{<auxiliary>}}, \\
w_{\text{normal}}, & \text{otherwise},
\end{cases}
\end{aligned}
\end{equation}
where $w_{\text{normal}} \leq 1 \leq w_{\text{aux}}$ assigns a higher weight to the sparse trigger token so the policy reliably acquires trigger timing without requiring a large learning rate that would distort general language representations.

\textbf{Stage 3: Joint Reasoning with Auxiliary Visuals.}
In Stage~3, after emitting \texttt{<auxiliary>}, the model completes Phase-2 reasoning conditioned on its \emph{self-generated} auxiliary tokens $A = V + g(D_\phi(H))$ under a joint objective:
\begin{equation}
  \mathcal{L}\;=\;\mathcal{L}_{\text{ntp}}\; +\; \lambda\,\mathcal{L}_{\text{vis}}.
\end{equation}
To smoothly bridge the training--inference distribution mismatch, we apply \textbf{teacher-forcing annealing}: Phase-2 initially conditions on ground-truth auxiliary features $E_{\mathrm{vis}}(I_{\text{aux}})$, and the probability of substituting self-generated $A$ is linearly annealed from $0 \to 1$ over training, gradually adapting the LLM to consume its own synthesized latents.

\subsection{ALPO: Auxiliary Latents Policy Optimization}

While three-stage SFT equips the model with foundational latent reasoning capabilities, supervised imitation alone has limited trajectory coverage. We therefore propose \textbf{ALPO} (Auxiliary Latents Policy Optimization) to jointly refine discrete reasoning and continuous auxiliary synthesis under outcome-driven reinforcement learning. Applying standard group-relative policy gradients (e.g., GRPO~\cite{guo2025deepseek,shao2024deepseekmath}) to SCALER faces a fundamental obstacle: the continuous auxiliary visual embeddings $e_\text{aux}\in\mathbb{R}^{N_v\times d}$ inserted between Phase~1 ($o_1$) and Phase~2 ($o_2$) are produced deterministically rather than sampled from a tractable token density. Naively caching $e_\text{aux}$ during rollout as detached fixed context severs the computation graph of $D_\phi$, preventing the detransformer from receiving task-level reward feedback and introducing context staleness as $\pi_\theta$ updates.

ALPO overcomes this bottleneck via \textbf{differentiable latent recomputation} during the actor update. During rollout, SCALER executes two-phase decoding under $\pi_{\theta_\text{old}}$ without retaining activation graphs. During the policy update on a sampled trajectory $o = o_1 \oplus o_2$, ALPO recomputes Phase-1 hidden states $H_\theta = f_\theta^{(\ell)}([X_v; q; o_1])$ under current parameters $\theta$ and regenerates $e_\text{aux}$ \emph{with full gradient tracking}:
\begin{equation}
  e_\text{aux}(\theta, \phi) = P\!\Big(V + g\big(D_\phi(H_\theta)\big)\Big),
\end{equation}
injecting it into the Phase-2 conditioning context to jointly optimize $(\theta, \phi)$ over discrete tokens $o_t \in o_1 \oplus o_2$:
\begin{equation}
  \begin{aligned}
  &\mathcal{J}_\text{ALPO}(\theta, \phi) = \\
  &\quad\mathbb{E}\!\left[\sum_t \min\!\left(\begin{aligned}
  &\tilde r_t(\theta,\phi)\hat{A}_t,\\
  &\;\mathrm{clip}(\tilde r_t(\theta,\phi),1-\epsilon,1+\epsilon)\hat{A}_t
  \end{aligned}\right)\right],
  \end{aligned}
\end{equation}
\begin{equation}
  \tilde r_t(\theta,\phi) = \frac{\pi_\theta\!\left(o_t\mid o_{<t},\;x,I,\;e_\text{aux}(\theta,\phi)\right)}{\pi_{\theta_\text{old}}\!\left(o_t\mid o_{<t},\;x,I,\;e_\text{aux}(\theta_\text{old},\phi_\text{old})\right)},
\end{equation}
where $\hat{A}_t$ denotes the group-normalized advantage across $G$ rollouts. Because $\log \pi_\theta(o_t \mid o_{<t}, x, I, e_\text{aux}(\theta,\phi))$ for Phase-2 tokens depends differentiably on $e_\text{aux}(\theta,\phi)$, policy gradients naturally backpropagate along the chain $\text{Phase-2 log-prob} \to e_\text{aux}(\theta,\phi) \to D_\phi \text{ and prefix LLM layers}$ without requiring an explicit density in continuous latent space.

\textbf{Reward Design.}
ALPO optimizes a composite reward:
\begin{equation}
  r = r_\text{acc} + \lambda_\text{fmt}\cdot r_\text{fmt} + \lambda_\text{aux}\cdot r_\text{aux},
\end{equation}
where $r_\text{acc}\in\{0,1\}$ verifies answer correctness, $r_\text{fmt}$ checks structural format validity, and $r_\text{aux} = \mathbb{I}[\text{auxiliary path executed} \wedge r_\text{acc} = 1]$ grants an auxiliary bonus \emph{if and only if} the model activates \texttt{<auxiliary>} and reaches the correct final answer. Conditioning $r_\text{aux}$ on answer correctness prevents degenerate trigger hacking and reinforces productive latent reasoning.

\section{Experiments}
\subsection{Experimental Settings}

We adopt Qwen2.5-VL-7B~\cite{bai2025qwen2} as the backbone with the vision encoder frozen (updating only the LLM and detransformer), trained on 210K SFT samples from Monet-SFT~\cite{wang2025monetreasoninglatentvisual} and VInteraction~\cite{qiao2025vthinker} plus 10K RL samples from ThymeRL~\cite{zhang2025thyme} and WeMath~\cite{qiao2024we} (setup details in Appendix~\ref{app:setup}). We evaluate on V*Bench~\cite{wu2024vstar}, HRBench4K/8K~\cite{wang2025divide_hrbench}, MME-RealWorld-Lite~\cite{zhang2024mme}, and OOD VisualPuzzles~\cite{song2025visualpuzzles} via \texttt{lmms-eval}~\cite{zhang2024lmmsevalrealitycheckevaluation} against textual reasoning (GPT-4o~\cite{gpt_4o}, GPT-6-luna~\cite{openai2026gpt6}, Qwen2.5-VL-7B~\cite{bai2025qwen2}, InternVL3-8B~\cite{zhu2025internvl3}, Pangea-7B~\cite{yue2024pangea}), tool-based (PixelReasoner~\cite{wang2025pixel}, DeepEyes~\cite{zheng2025deepeyes}, DeepEyesV2~\cite{hong2026deepeyesv2}), and visual latent methods (LVR~\citep{li2025latent}, Monet~\cite{wang2025monetreasoninglatentvisual}, ILVR~\cite{dong2025ilvr}, HyLaR~\cite{cheng2026hybrid}, LASER~\cite{wang2026laser}, UniVLR~\cite{jiang2026univlr}, LUT-7B~\cite{kang2026lut}).

\subsection{Main Results}
\begin{table}[!t]
    \centering
    \caption{\textbf{Performance on VisualPuzzles (OOD).} Best open-source results are in \textbf{bold}, and second-best are \underline{underlined}.}
    \label{tab:exp-visualpuzzles}
    \renewcommand{\arraystretch}{1.05}
    \setlength{\tabcolsep}{3.5pt}
    \resizebox{\linewidth}{!}{%
    \begin{tabular}{lcccccc}
        \toprule
        \multirow{2}{*}{\textbf{Methods}} & \multicolumn{6}{c}{\textbf{VisualPuzzles}}\\
        \cmidrule(lr){2-7}
        & Overall & Algo. & Anal. & Dedu. & Indu. & Spat. \\
        \midrule
        \rowcolor[HTML]{EBEBFF}
        \multicolumn{7}{c}{\textit{Textual Reasoning}} \\
        GPT-4o$_{\text{(2024)}}$~\cite{gpt_4o} & 41.3* & 49.2* & 58.3* & 49.0* & 27.3* & 26.2* \\
        GPT-6-luna$_{\text{(2026)}}$~\cite{openai2026gpt6} & 41.52 & 56.49 & 34.60 & 55.00 & 24.40 & 36.01 \\
        Qwen2.5-VL-7B$_{\text{(2025)}}$~\cite{bai2025qwen2} & 32.71 & 37.02 & 21.80 & \underline{47.50} & 26.32 & 21.80 \\
        Pangea-7B$_{\text{(2024)}}$~\citep{yue2024pangea} & 31.3* & 32.4* & 23.7* & 38.5* & 28.7* & \textbf{32.5*} \\
        \midrule
        \rowcolor[HTML]{F9E6EC}
        \multicolumn{7}{c}{\textit{Tool-based Visual Reasoning}} \\
        DeepEyes$_{\text{(2026)}}$~\citep{zheng2025deepeyes} & 32.96 & 37.79 & 27.01 & 41.00 & 26.79 & 27.01 \\
        DeepEyesV2$_{\text{(2026)}}$~\citep{hong2026deepeyesv2} & 34.66 & \textbf{41.73} & \underline{32.23} & 45.13 & 25.84 & 29.47 \\
        \midrule
        \rowcolor[HTML]{E6FFE6}
        \multicolumn{7}{c}{\textit{Visual Latent Reasoning}} \\
        LVR$_{\text{(2025)}}$~\citep{li2025latent} & 28.51 & 29.39 & 26.54 & 40.00 & 25.84 & 23.08 \\
        Monet$_{\text{(2026)}}$~\citep{wang2025monetreasoninglatentvisual} & 28.34 & 26.34 & 28.44 & 42.00 & 22.01 & 25.17 \\
        HyLaR$_{\text{(2026)}}$~\citep{cheng2026hybrid} & 30.05 & 35.11 & 29.38 & 37.50 & 24.40 & 24.83 \\
        LASER$_{\text{(2026)}}$~\citep{wang2026laser} & \underline{35.10} & 37.79 & 29.38 & \textbf{48.00} & 29.19 & \underline{32.17} \\
        UniVLR$_{\text{(2026)}}$~\citep{jiang2026univlr} & 31.25 & 34.73 & 26.54 & 38.00 & \textbf{30.62} & 27.27 \\
        \textbf{SCALER-SFT (Ours)} & 34.16 & \underline{40.08} & 30.81 & 43.50 & \underline{30.14} & 27.62 \\
        \textbf{SCALER-7B (Ours)} & \textbf{35.27} & 38.93 & \textbf{34.60} & 40.50 & \textbf{30.62} & \underline{32.17} \\
        \rowcolor[HTML]{ECECEC}
        \textit{$\Delta$} (vs Qwen2.5-VL-7B) & \textcolor[rgb]{0,0.6,0}{+2.56} & \textcolor[rgb]{0,0.6,0}{+1.91} & \textcolor[rgb]{0,0.6,0}{+12.80} & \textcolor[rgb]{0.6,0,0}{-7.00} & \textcolor[rgb]{0,0.6,0}{+4.30} & \textcolor[rgb]{0,0.6,0}{+10.37} \\
        \bottomrule
    \end{tabular}%
    }
    \vspace{-1em}
\end{table}

\paragraph{Overall Performance.}
As shown in Table~\ref{tab:exp-main}, SCALER-7B achieves state-of-the-art open-source accuracy on three of the four perception and real-world reasoning benchmarks (ranking a close second on HRBench-8K Overall while leading FCP by +10.0\% over HyLaR), consistently outperforming tool-based and visual latent baselines (Appendix~\ref{app:ilvr_comparison} for ILVR~\cite{dong2025ilvr}). Two core mechanisms drive these gains:

\textbf{(1) Single-pass context-anchored generation preserves fine-grained spatial and cross-instance structure.} SCALER's advantage is most pronounced on tasks demanding dense spatial localization and multi-object relational reasoning, such as MME-RealWorld-Lite Reasoning (\textbf{55.73\%}, \textbf{+16.00\%} over the backbone), HRBench8K FCP, and V*\,Bench Spatial. Because the detransformer synthesizes spatially aligned auxiliary tokens in a single forward pass from full-sequence hidden states and residually fuses them with $V$ ($A = V + g(A_0)$), each token independently reconstructs local visual evidence within global scene context without autoregressive erosion.

\textbf{(2) Non-saturating visual latents enable a 7B model to surpass frontier proprietary systems.} By scaling latent sequence length without representation saturation, SCALER-7B substantially outperforms both GPT-4o and GPT-6-luna~\cite{openai2026gpt6} on fine-grained perception (reaching \textbf{84.29\%} on V*\,Bench, \textbf{+9.42\%} over GPT-6-luna) and complex real-world reasoning (reaching \textbf{57.84\%} on MME-RealWorld-Lite, \textbf{+10.47\%} over GPT-6-luna), demonstrating that high-capacity visual latents can bridge the model-scale gap.

\paragraph{Out-of-Distribution Generalization.}
On OOD VisualPuzzles (Table~\ref{tab:exp-visualpuzzles}), an abstract visual logic benchmark entirely absent from our training distribution, SCALER-7B achieves the highest overall accuracy (\textbf{35.27\%}) among open-source models. Gains over the backbone are driven by sub-tasks requiring relational and structural abstraction, including \emph{analogical} (\textbf{+12.80\%}, matching GPT-6-luna~\cite{openai2026gpt6}), \emph{spatial} (\textbf{+10.37\%}, surpassing GPT-4o), and \emph{inductive} (\textbf{+4.30\%}, surpassing GPT-4o and GPT-6-luna) reasoning, while SCALER-SFT leads all latent reasoning models on \emph{algorithmic} reasoning. Predicting context-anchored residuals ($\Delta_{\text{aux}}$) thus equips the detransformer with transferable spatial operations.

\subsection{Analysis of the Latent Scaling Bottleneck}
\label{sec:scaling_bottleneck}
\begin{figure*}[!t]
  \centering
  \includegraphics[width=1.0\linewidth]{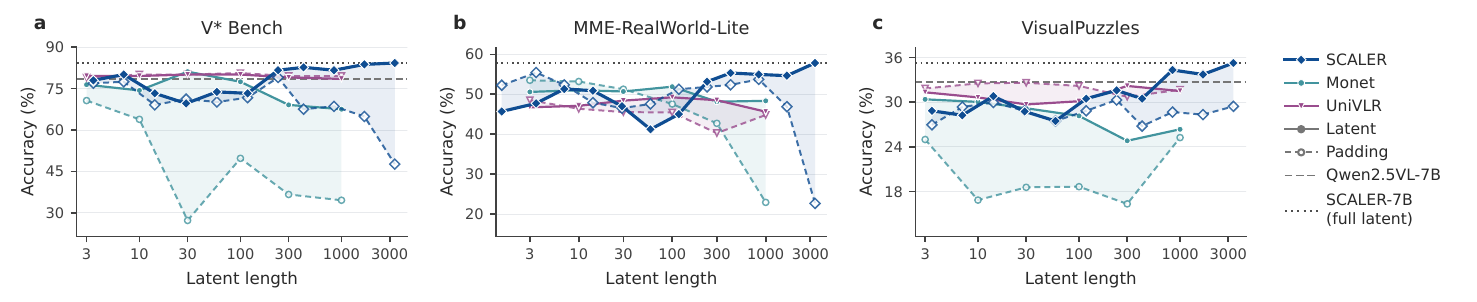}
  \caption{\textbf{Latent Length Scaling \& Meaningless Padding Experiments on (a) V*\,Bench, (b) MME-RealWorld-Lite, and (c) VisualPuzzles.}
      Solid lines: normal inference across latent lengths; dashed lines: replacing generated latents with meaningless padding; shaded regions: contribution of genuine latent semantics.}
    \label{fig:filling}
\end{figure*}

To examine how latent length interacts with task difficulty, we evaluate Monet~\cite{wang2025monetreasoninglatentvisual}, LVR~\cite{li2025latent}, LASER~\cite{wang2026laser}, UniVLR~\cite{jiang2026univlr}, and SCALER across budgets $K \in \{0, 3, 10, 30, 100, 300, 1000\}$ (Figure~\ref{fig:confusion_matrix}). All 4,878 test samples across the five benchmarks are stratified into five complexity tiers (C1--C5) by DeepSeek-v4.1-flash~\cite{deepseekai2026deepseekv41flashpushinglimitskv} based on perceptual granularity and reasoning hops (Appendix~\ref{app:complexity_classification}), revealing three key phenomena:

\textbf{(1) Shallow scaling at low complexity ($K \leq 30$, C1--C2).} Moving from $K=0$ to $K \in \{3, 10, 30\}$ improves all existing methods on C1--C2, reconciling prior observations that a few latent tokens suffice to provide coarse attentional cues for simple single-hop perception.

\textbf{(2) The Latent Scaling Bottleneck at high complexity ($K \geq 100$, C3--C5).} In the upper-right regime ($K \geq 100$, C3--C5) where multi-step reasoning is needed most, LVR stagnates while Monet, LASER, and UniVLR systematically degrade, as autoregressive chains accumulate saturated tokens that act as context interference.

\textbf{(3) SCALER breaks the bottleneck across complexity and length.} SCALER (Figure~\ref{fig:confusion_matrix}(e)) achieves overall scaling gains on C1--C2 (after a brief pooling dip at $K=30$) and unlocks substantial gains on C4--C5 at $K \in \{100, 300, 1000\}$ by preserving fine spatial resolution.

\subsection{Latent Length Scaling and Meaningless Padding Experiments}
\label{sec:padding_exp}
To test whether generated latents carry genuine visual semantics across lengths, we evaluate length scaling and meaningless padding on V*\,Bench, MME-RealWorld-Lite, and VisualPuzzles from $K=3$ to full length ($\sim$3000+ tokens; Figure~\ref{fig:filling}). For autoregressive baselines, we vary the number of generated tokens; for SCALER, we control sequence length via spatial pooling on the detransformer output. In the \textbf{meaningless padding intervention}, generated latents are replaced at inference with fixed padding embeddings of identical length. Additional baselines in Appendix~\ref{app:padding_supp} and qualitative case and attention analyses in Appendices~\ref{app:case_study}--\ref{app:attention_vis} further show that longer budgets shift visual attention from salient distractors onto fine-grained targets.

\paragraph{Length Scaling Patterns (Solid Lines).} Across all three benchmarks, as latent length increases from 3 to 1000 (Figures~\ref{fig:motivation}(a) and~\ref{fig:filling}(a--c)), autoregressive methods either systematically degrade (Monet) or remain largely flat and slightly decline at $K=1000$ (UniVLR), failing to extract additional visual evidence from longer chains. By contrast, SCALER consistently exhibits a distinctive \textbf{three-phase trajectory} across V*\,Bench, MME-RealWorld-Lite, and VisualPuzzles: under extreme pooling ($K \leq 10$), the model relies primarily on textual reasoning with minimal visual distortion; under moderate pooling ($K \approx 15\text{--}60$), accuracy dips (especially at $K \approx 30\text{--}60$) because the auxiliary sequence is long enough to shift attention but too heavily compressed to preserve spatial layout; under weak-to-no pooling ($K \geq 100$ to full resolution), accuracy climbs steadily to its global peak across all three tasks. This universal trajectory confirms a critical mechanism: \emph{excessively pooled ($\geq$128$\times$) visual features lose spatial fidelity and act as noisy distractors}, whereas restoring spatial resolution unlocks sustained scaling.

\paragraph{Meaningless Padding Analysis (Dashed Lines).} Replacing latent tokens with meaningless padding directly probes their causal semantic contribution across all three benchmarks (Figure~\ref{fig:filling}(a--c)), revealing three distinct regimes: (i)~for UniVLR, the dashed padding curve stays close to or overlaps with normal inference across lengths on V*\,Bench, MME-RealWorld-Lite, and VisualPuzzles, indicating that its decoder largely bypasses the semantic content of generated latent tokens; (ii)~for Monet, meaningless padding causes a sharp accuracy drop (plunging at $K \geq 30$ on V*\,Bench and VisualPuzzles and at $K \geq 300$ on MME-RealWorld-Lite), showing strong structural coupling to the latent distribution, yet combined with its stagnating or declining solid curve, this coupling turns accumulating representation saturation into destructive interference over long chains; and (iii)~for SCALER, solid and dashed curves remain close under heavy pooling ($K < 100$, confirming semantic loss under over-compression), but diverge sharply as resolution recovers ($K \geq 100$ to $> 3000$ tokens, where padding collapses while normal inference peaks), proving that longer SCALER sequences contribute indispensable visual semantics.

\subsection{Analysis of Step-wise Cosine Novelty and Latent Representation Saturation}
\label{sec:novelty_exp}

To explain \emph{why} autoregressive latents fail to scale, we quantify each token's directional novelty relative to all preceding tokens via \textbf{Step-wise Cosine Novelty}:
\begin{equation}
  \mathrm{novelty}_t = 1 - \max_{s < t} \cos(e_t, e_s) = \min_{s < t}\bigl[1 - \cos(e_t, e_s)\bigr],
\end{equation}
where $\cos(e_t, e_s) = \frac{\langle e_t, e_s \rangle}{\|e_t\|_2 \|e_s\|_2}$ for $t \ge 2$. Small $\mathrm{novelty}_t$ indicates near-duplication of earlier tokens (\textbf{Latent Representation Saturation}). As shown in Figure~\ref{fig:motivation}(b) on MME-RealWorld-Lite, both Monet and UniVLR decay exponentially within the first $\sim$30 tokens and saturate at $10^{-3}\text{--}10^{-4}$ as recursive conditioning traps the generator in a narrow subspace. In contrast, SCALER sustains high novelty (\textbf{$10^{-1}\text{--}10^{0}$}) across all 1000 tokens (two to three orders of magnitude above baselines) by generating all tokens in a single pass anchored to distinct spatial regions.

\subsection{Ablation Studies}

Table~\ref{tab:exp-ablation} validates four core design choices:

\textbf{(1) Downstream reasoning genuinely relies on edited auxiliary visual semantics.} Disabling auxiliary generation noticeably degrades accuracy (e.g., V*\,Bench drops from 84.29\% to 78.01\%), while replacing $e_\text{aux}$ with meaningless placeholders triggers catastrophic collapse across benchmarks (down to 47.64\% on V*\,Bench), proving that Phase-2 reasoning actively attends to the semantic content of $e_\text{aux}$ rather than treating it as a positional bias. Replacing $e_\text{aux}$ with raw visual features $V$ also consistently hurts accuracy, confirming the necessity of query-conditioned visual editing.

\textbf{(2) Stage~1 detransformer pretraining is an essential prerequisite.} Removing Stage~1 causes severe collapse across all benchmarks (e.g., 47.64\% on V*\,Bench and 18.49\% on VisualPuzzles), as an unaligned detransformer injects incoherent perturbations that corrupt the LLM's visual context during joint training.

\begin{table}[!t]
    \centering
    \caption{\textbf{Ablation study.} Top block: inference-time interventions on SCALER-7B; bottom block: RL and SFT stage ablations relative to SCALER-SFT. Overall accuracy (\%) is reported.}
    \label{tab:exp-ablation}
    \renewcommand{\arraystretch}{1.10}
    \setlength{\tabcolsep}{3.5pt}
    \resizebox{\linewidth}{!}{%
    \begin{tabular}{lccccc}
        \toprule
        \textbf{Methods} & \textbf{V$^*$} & \textbf{HR-4K} & \textbf{HR-8K} & \textbf{MME-RW} & \textbf{VP} \\
        \midrule
        \multicolumn{6}{l}{\textit{Full Pipeline \& Inference-Time Interventions}} \\
        \rowcolor[HTML]{EBEBFF}
        \textbf{SCALER-7B (Ours, full)} & \textbf{84.29} & \textbf{75.25} & \textbf{68.88} & \textbf{57.84} & \textbf{35.27} \\
        \quad + disable auxiliary & 78.01 & 68.88 & 62.88 & 51.22 & 31.16 \\
        \quad + replace w/ placeholders & 47.64 & 40.12 & 42.62 & 36.84 & 29.45 \\
        \quad + replace w/ raw visual emb. & 80.10 & 71.25 & 67.38 & 53.99 & 31.93 \\
        \midrule
        \multicolumn{6}{l}{\textit{RL \& SFT Stage Ablations}} \\
        \rowcolor[HTML]{F9E6EC}
        \textbf{SCALER-SFT (w/o ALPO)} & \underline{82.72} & 74.25 & 66.75 & 52.84 & \underline{34.16} \\
        \quad + GRPO (replace ALPO) & 82.20 & \textbf{75.25} & 67.62 & \underline{55.13} & 32.88 \\
        \quad w/o Stage 1 (pretraining) & 47.64 & 45.62 & 48.00 & 38.67 & 18.49 \\
        \quad w/o token-level loss (Stage 2) & 81.68 & 72.25 & \underline{67.75} & 50.39 & 33.05 \\
        \quad w/o teacher forcing (Stage 3) & 80.10 & 71.00 & 66.50 & 48.57 & 33.48 \\
        \quad w/o Stage 3 & 77.49 & 70.75 & 65.62 & 41.90 & 29.88 \\
        \bottomrule
    \end{tabular}%
    }
    \vspace{-1em}
\end{table}

\textbf{(3) Stage~2 trigger weighting and Stage~3 annealing resolve complementary bottlenecks.} Removing Stage~2 loss weighting weakens trigger learning on hard samples, while removing Stage~3 teacher-forcing annealing (or Stage~3 entirely) degrades multi-step reasoning, confirming the necessity of gradual adaptation to self-generated latents.

\textbf{(4) ALPO outperforms GRPO via differentiable latent recomputation.} ALPO surpasses SCALER-SFT across all five benchmarks and outperforms standard GRPO on four out of five benchmarks (tying on HRBench-4K), showing that gradient-preserving latent recomputation with correctness-conditioned rewards effectively propagates RL signals into continuous latent synthesis.

\section{Conclusion}

We uncover the \textbf{Latent Scaling Bottleneck}, where existing visual latent reasoning methods scale only within short budgets ($\leq$30 tokens) on simple tasks yet stagnate or degrade on complex tasks and longer sequences. Controlled diagnostics trace this bottleneck to \textbf{Latent Representation Saturation} under autoregressive conditioning and \textbf{Over-Compressed Supervision Targets} from heavily pooled ($\geq$128$\times$) embeddings. To break this bottleneck, we propose \textbf{SCALER}, whose lightweight detransformer synthesizes spatially aligned auxiliary tokens from full-sequence hidden states in a single forward pass. Our experiments form a closed diagnostic loop: preserving spatial resolution overcomes the compression dip to unlock scaling on complex tasks; single-pass context-anchored synthesis prevents representation saturation across 1000+ tokens; and three-stage SFT with \textbf{ALPO} differentiable latent recomputation grounds multi-step reasoning in synthesized semantics, achieving state-of-the-art open-source accuracy across benchmarks.

\section*{LLM Usage Statement}
GPT-4o is used for CoT distillation in training data construction (Appendix~\ref{app:sft_data}). Qwen3-VL-235B-A22B serves as a judge in the preliminary architecture study (Appendix~\ref{app:layer_selection}), and DeepSeek-v4.1-flash is used to stratify evaluation samples into task complexity levels C1--C5 (Appendix~\ref{app:complexity_classification}). Additionally, LLMs are used for paper writing polish and non-core code implementation; all LLM-generated content beyond data production and evaluation is manually reviewed.

{
  \small
  \bibliographystyle{ieeenat_fullname}
  \bibliography{reference_cvpr}

@inproceedings{xu2025llava,
  title={Llava-cot: Let vision language models reason step-by-step},
  author={Xu, Guowei and Jin, Peng and Wu, Ziang and Li, Hao and Song, Yibing and Sun, Lichao and Yuan, Li},
  booktitle={Proceedings of the IEEE/CVF International Conference on Computer Vision},
  pages={2087--2098},
  year={2025}
}

@inproceedings{zhang2024lmmsevalrealitycheckevaluation,
   title={LMMs-Eval: Reality Check on the Evaluation of Large Multimodal Models},
   url={http://dx.doi.org/10.18653/v1/2025.findings-naacl.51},
   DOI={10.18653/v1/2025.findings-naacl.51},
   booktitle={Findings of the Association for Computational Linguistics: NAACL 2025},
   publisher={Association for Computational Linguistics},
   author={Zhang, Kaichen and Li, Bo and Zhang, Peiyuan and Pu, Fanyi and Cahyono, Joshua Adrian and Hu, Kairui and Liu, Shuai and Zhang, Yuanhan and Yang, Jingkang and Li, Chunyuan and Liu, Ziwei},
   year={2025},
   pages={881--916} }

@article{chen2025towards,
  title={Towards reasoning era: A survey of long chain-of-thought for reasoning large language models},
  author={Chen, Qiguang and Qin, Libo and Liu, Jinhao and Peng, Dengyun and Guan, Jiannan and Wang, Peng and Hu, Mengkang and Zhou, Yuhang and Gao, Te and Che, Wanxiang},
  journal={arXiv preprint arXiv:2503.09567},
  year={2025}
}

@inproceedings{zheng2025deepeyes,
  title={DeepEyes: Incentivizing ``Thinking with Images'' via Reinforcement Learning},
  author={Zheng, Ziwei and Yang, Michael and Hong, Jack and Zhao, Chenxiao and Xu, Guohai and Yang, Le and Shen, Chao and Yu, Xing},
  booktitle={International Conference on Learning Representations},
  pages={126775--126798},
  year={2026}
}

@article{bai2025qwen2,
  title={{Qwen2.5-VL Technical Report}},
  author={Bai, Shuai and Chen, Keqin and Liu, Xuejing and Wang, Jialin and Ge, Wenbin and Song, Sibo and Dang, Kai and Wang, Peng and Wang, Shijie and Tang, Jun and others},
  journal={arXiv preprint arXiv:2502.13923},
  year={2025}
}

@article{jaech2024openai,
  title={Openai o1 system card},
  author={Jaech, Aaron and Kalai, Adam and Lerer, Adam and Richardson, Adam and El-Kishky, Ahmed and Low, Aiden and Helyar, Alec and Madry, Aleksander and Beutel, Alex and Carney, Alex and others},
  journal={arXiv preprint arXiv:2412.16720},
  year={2024}
}

@article{guo2025deepseek,
  title={Deepseek-r1: Incentivizing reasoning capability in llms via reinforcement learning},
  author={Guo, Daya and Yang, Dejian and Zhang, Haowei and Song, Junxiao and Wang, Peiyi and Zhu, Qihao and Xu, Runxin and Zhang, Ruoyu and Ma, Shirong and Bi, Xiao and others},
  journal={arXiv preprint arXiv:2501.12948},
  year={2025}
}

@article{shao2024deepseekmath,
  title={Deepseekmath: Pushing the limits of mathematical reasoning in open language models},
  author={Shao, Zhihong and Wang, Peiyi and Zhu, Qihao and Xu, Runxin and Song, Junxiao and Bi, Xiao and Zhang, Haowei and Zhang, Mingchuan and Li, YK and Wu, Yang and others},
  journal={arXiv preprint arXiv:2402.03300},
  year={2024}
}

@article{jian2025look,
  title={Look again, think slowly: Enhancing visual reflection in vision-language models},
  author={Jian, Pu and Wu, Junhong and Sun, Wei and Wang, Chen and Ren, Shuo and Zhang, Jiajun},
  journal={arXiv preprint arXiv:2509.12132},
  year={2025}
}

@article{loshchilov2017decoupled,
  title={Decoupled weight decay regularization},
  author={Loshchilov, Ilya and Hutter, Frank},
  journal={arXiv preprint arXiv:1711.05101},
  year={2017}
}

@article{chen2025mindgpt,
  title={Mindgpt-4ov: An enhanced mllm via a multi-stage post-training paradigm},
  author={Chen, Wei and Du, Chaoqun and Gu, Feng and He, Wei and Li, Qizhen and Liu, Zide and Pan, Xuhao and Ren, Chang and Rao, Xudong and Wang, Chenfeng and others},
  journal={arXiv preprint arXiv:2512.02895},
  year={2025}
}

@article{xu2025visualplanningletsthink,
      title={Visual Planning: Let's Think Only with Images}, 
      author={Yi Xu and Chengzu Li and Han Zhou and Xingchen Wan and Caiqi Zhang and Anna Korhonen and Ivan Vuli{\'c}},
      journal={arXiv preprint arXiv:2505.11409},
      year={2026},
      eprint={2505.11409},
      archivePrefix={arXiv},
      primaryClass={cs.LG},
      url={https://arxiv.org/abs/2505.11409}, 
}

@article{mavors,
  title={Mavors: Multi-granularity video representation for multimodal large language model},
  author={Shi, Yang and Liu, Jiaheng and Guan, Yushuo and Wu, Zhenhua and Zhang, Yuanxing and Wang, Zihao and Lin, Weihong and Hua, Jingyun and Wang, Zekun and Chen, Xinlong and others},
  journal={arXiv preprint arXiv:2504.10068},
  year={2025}
}

@article{yeo2025demystifying,
  title={Demystifying long chain-of-thought reasoning in llms},
  author={Yeo, Edward and Tong, Yuxuan and Niu, Morry and Neubig, Graham and Yue, Xiang},
  journal={arXiv preprint arXiv:2502.03373},
  year={2025}
}

@article{wang2025internvl3,
  title={Internvl3. 5: Advancing open-source multimodal models in versatility, reasoning, and efficiency},
  author={Wang, Weiyun and Gao, Zhangwei and Gu, Lixin and Pu, Hengjun and Cui, Long and Wei, Xingguang and Liu, Zhaoyang and Jing, Linglin and Ye, Shenglong and Shao, Jie and others},
  journal={arXiv preprint arXiv:2508.18265},
  year={2025}
}

@article{liu2023visual,
  title={Visual instruction tuning},
  author={Liu, Haotian and Li, Chunyuan and Wu, Qingyang and Lee, Yong Jae},
  journal={Advances in neural information processing systems},
  volume={36},
  pages={34892--34916},
  year={2023}
}

@article{peng2025lmm,
  title={Lmm-r1: Empowering 3b lmms with strong reasoning abilities through two-stage rule-based rl},
  author={Peng, Yingzhe and Zhang, Gongrui and Zhang, Miaosen and You, Zhiyuan and Liu, Jie and Zhu, Qipeng and Yang, Kai and Xu, Xingzhong and Geng, Xin and Yang, Xu},
  journal={arXiv preprint arXiv:2503.07536},
  year={2025}
}

@article{wei2025advancing,
  title={Advancing multimodal reasoning via reinforcement learning with cold start},
  author={Wei, Lai and Li, Yuting and Zheng, Kaipeng and Wang, Chen and Wang, Yue and Kong, Linghe and Sun, Lichao and Huang, Weiran},
  journal={arXiv preprint arXiv:2505.22334},
  year={2025}
}

@article{jiang2025vlm,
  title={VLM-R$^3$: Region Recognition, Reasoning, and Refinement for Enhanced Multimodal Chain-of-Thought},
  author={Jiang, Chaoya and Heng, Yongrui and Ye, Wei and Yang, Han and Xu, Haiyang and Yan, Ming and Zhang, Ji and Huang, Fei and Zhang, Shikun},
  journal={arXiv preprint arXiv:2505.16192},
  year={2025}
}

@article{li2024llava,
  title={Llava-onevision: Easy visual task transfer},
  author={Li, Bo and Zhang, Yuanhan and Guo, Dong and Zhang, Renrui and Li, Feng and Zhang, Hao and Zhang, Kaichen and Zhang, Peiyuan and Li, Yanwei and Liu, Ziwei and others},
  journal={arXiv preprint arXiv:2408.03326},
  year={2024}
}

@article{comanici2025gemini,
  title={Gemini 2.5: Pushing the frontier with advanced reasoning, multimodality, long context, and next generation agentic capabilities},
  author={Comanici, Gheorghe and Bieber, Eric and Schaekermann, Mike and Pasupat, Ice and Sachdeva, Noveen and Dhillon, Inderjit and Blistein, Marcel and Ram, Ori and Zhang, Dan and Rosen, Evan and others},
  journal={arXiv preprint arXiv:2507.06261},
  year={2025}
}

@article{yu2025perception,
  title={Perception-r1: Pioneering perception policy with reinforcement learning},
  author={Yu, En and Lin, Kangheng and Zhao, Liang and Yin, Jisheng and Wei, Yana and Peng, Yuang and Wei, Haoran and Sun, Jianjian and Han, Chunrui and Ge, Zheng and others},
  journal={arXiv preprint arXiv:2504.07954},
  year={2025}
}

@article{yang2025mirage,
  title={Machine Mental Imagery: Empower Multimodal Reasoning with Latent Visual Tokens},
  author={Yang, Zeyuan and Yu, Xueyang and Chen, Delin and Shen, Maohao and Gan, Chuang},
  journal={arXiv preprint arXiv:2506.17218},
  year={2025}
}

@article{wang2025visualprm,
  title={Visualprm: An effective process reward model for multimodal reasoning},
  author={Wang, Weiyun and Gao, Zhangwei and Chen, Lianjie and Chen, Zhe and Zhu, Jinguo and Zhao, Xiangyu and Liu, Yangzhou and Cao, Yue and Ye, Shenglong and Zhu, Xizhou and others},
  journal={arXiv preprint arXiv:2503.10291},
  year={2025}
}

@article{chen2026streamingclaw,
  title={StreamingClaw Technical Report},
  author={Chen, Jiawei and Chen, Zhe and Du, Chaoqun and He, Maokui and He, Wei and Li, Hengtao and Li, Qizhen and Liu, Zide and Ma, Hao and Pan, Xuhao and others},
  journal={arXiv preprint arXiv:2603.22120},
  year={2026}
}

@article{li2025latent,
  title={Latent visual reasoning},
  author={Li, Bangzheng and Sun, Ximeng and Liu, Jiang and Wang, Ze and Wu, Jialian and Yu, Xiaodong and Chen, Hao and Barsoum, Emad and Chen, Muhao and Liu, Zicheng},
  journal={arXiv preprint arXiv:2509.24251},
  year={2025}
}

@article{wang2025pixel,
  title={Pixel reasoner: Incentivizing pixel-space reasoning with curiosity-driven reinforcement learning},
  author={Wang, Haozhe and Su, Alex and Ren, Weiming and Lin, Fangzhen and Chen, Wenhu},
  journal={arXiv preprint arXiv:2505.15966},
  year={2025}
}

@article{gpt_4o,
  title={Gpt-4o system card},
  author={Hurst, Aaron and Lerer, Adam and Goucher, Adam P and Perelman, Adam and Ramesh, Aditya and Clark, Aidan and Ostrow, AJ and Welihinda, Akila and Hayes, Alan and Radford, Alec and others},
  journal={arXiv preprint arXiv:2410.21276},
  year={2024}
}

@article{zhang2025thyme,
  title={Thyme: Think Beyond Images},
  author={Zhang, Yi-Fan and Lu, Xingyu and Yin, Shukang and Fu, Chaoyou and Chen, Wei and Hu, Xiao and Wen, Bin and Jiang, Kaiyu and Liu, Changyi and Zhang, Tianke and others},
  journal={arXiv preprint arXiv:2508.11630},
  year={2025}
}

@article{zhang2025adaptive,
  title={Adaptive Chain-of-Focus Reasoning via Dynamic Visual Search and Zooming for Efficient VLMs},
  author={Zhang, Xintong and Gao, Zhi and Zhang, Bofei and Li, Pengxiang and Zhang, Xiaowen and Liu, Yang and Yuan, Tao and Wu, Yuwei and Jia, Yunde and Zhu, Song-Chun and others},
  journal={arXiv preprint arXiv:2505.15436},
  year={2025}
}

@inproceedings{wang2025monetreasoninglatentvisual,
      title={Monet: Reasoning in Latent Visual Space Beyond Images and Language}, 
      author={Qixun Wang and Yang Shi and Yifei Wang and Yuanxing Zhang and Pengfei Wan and Kun Gai and Xianghua Ying and Yisen Wang},
      year={2026},
      booktitle={CVPR}
}

@article{huang2025visualtoolagent,
  title={Visualtoolagent (vista): A reinforcement learning framework for visual tool selection},
  author={Huang, Zeyi and Ji, Yuyang and Rajan, Anirudh Sundara and Cai, Zefan and Xiao, Wen and Wang, Haohan and Hu, Junjie and Lee, Yong Jae},
  journal={arXiv preprint arXiv:2505.20289},
  year={2025}
}

@article{coconut,
  title={Training Large Language Models to Reason in a Continuous Latent Space},
  author={Hao, Shibo and Sukhbaatar, Sainbayar and Su, DiJia and Li, Xian and Hu, Zhiting and Weston, Jason and Tian, Yuandong},
  journal={arXiv preprint arXiv:2412.06769},
  year={2024}
}

@article{he2025diffthinkergenerativemultimodalreasoning,
      title={DiffThinker: Towards Generative Multimodal Reasoning with Diffusion Models}, 
      author={Zefeng He and Xiaoye Qu and Yafu Li and Tong Zhu and Siyuan Huang and Yu Cheng},
      journal={arXiv preprint arXiv:2512.24165},
      year={2025},
      eprint={2512.24165},
      archivePrefix={arXiv},
      primaryClass={cs.CV},
      url={https://arxiv.org/abs/2512.24165}, 
}

@article{wei2025sim,
  title={SIM-CoT: Supervised Implicit Chain-of-Thought},
  author={Wei, Xilin and Liu, Xiaoran and Zang, Yuhang and Dong, Xiaoyi and Cao, Yuhang and Wang, Jiaqi and Qiu, Xipeng and Lin, Dahua},
  journal={arXiv preprint arXiv:2509.20317},
  year={2025}
}

@article{wang2025synadapt,
  title={Synadapt: Learning adaptive reasoning in large language models via synthetic continuous chain-of-thought},
  author={Wang, Jianwei and Wu, Ziming and Lai, Fuming and Lian, Shaobing and Zeng, Ziqian},
  journal={arXiv preprint arXiv:2508.00574},
  year={2025}
}

@article{vaswani2017attention,
  title={Attention is all you need},
  author={Vaswani, Ashish and Shazeer, Noam and Parmar, Niki and Uszkoreit, Jakob and Jones, Llion and Gomez, Aidan N and Kaiser, {\L}ukasz and Polosukhin, Illia},
  journal={Advances in neural information processing systems},
  volume={30},
  year={2017}
}

@article{cheng2025visual,
  title={Visual thoughts: A unified perspective of understanding multimodal chain-of-thought},
  author={Cheng, Zihui and Chen, Qiguang and Xu, Xiao and Wang, Jiaqi and Wang, Weiyun and Fei, Hao and Wang, Yidong and Wang, Alex Jinpeng and Chen, Zhi and Che, Wanxiang and others},
  journal={arXiv preprint arXiv:2505.15510},
  year={2025}
}

@inproceedings{lowe1999object,
  title={Object recognition from local scale-invariant features},
  author={Lowe, David G},
  booktitle={Proceedings of the seventh IEEE international conference on computer vision},
  volume={2},
  pages={1150--1157},
  year={1999},
  organization={Ieee}
}

@article{lowe2004distinctive,
  title={Distinctive image features from scale-invariant keypoints},
  author={Lowe, David G},
  journal={International journal of computer vision},
  volume={60},
  number={2},
  pages={91--110},
  year={2004},
  publisher={Springer}
}

@article{chen2025learning,
  title={Learning Only with Images: Visual Reinforcement Learning with Reasoning, Rendering, and Visual Feedback},
  author={Chen, Yang and Shen, Yufan and Huang, Wenxuan and Zhou, Shen and Lin, Qunshu and Cai, Xinyu and Yu, Zhi and Shi, Botian and Qiao, Yu},
  journal={arXiv preprint arXiv:2507.20766},
  year={2025}
}

@inproceedings{fu2025refocus,
  title={Refocus: Visual editing as a chain of thought for structured image understanding},
  author={Fu, Xingyu and Liu, Minqian and Yang, Zhengyuan and Corring, John and Lu, Yijuan and Yang, Jianwei and Roth, Dan and Florencio, Dinei and Zhang, Cha},
  booktitle={ICML},
  year={2025}
}

@article{hu2024visual,
  title={Visual sketchpad: Sketching as a visual chain of thought for multimodal language models},
  author={Hu, Yushi and Shi, Weijia and Fu, Xingyu and Roth, Dan and Ostendorf, Mari and Zettlemoyer, Luke and Smith, Noah A and Krishna, Ranjay},
  journal={Advances in Neural Information Processing Systems},
  year={2024}
}

@article{su2025openthinkimg,
  title={Openthinkimg: Learning to think with images via visual tool reinforcement learning},
  author={Su, Zhaochen and Li, Linjie and Song, Mingyang and Hao, Yunzhuo and Yang, Zhengyuan and Zhang, Jun and Chen, Guanjie and Gu, Jiawei and Li, Juntao and Qu, Xiaoye and others},
  journal={arXiv preprint arXiv:2505.08617},
  year={2025}
}

@inproceedings{qi2024cogcom,
  title={Cogcom: A visual language model with chain-of-manipulations reasoning},
  author={Qi, Ji and Ding, Ming and Wang, Weihan and Bai, Yushi and Lv, Qingsong and Hong, Wenyi and Xu, Bin and Hou, Lei and Li, Juanzi and Dong, Yuxiao and others},
  booktitle={ICLR},
  year={2025}
}

@article{li2025imagine,
  title={Imagine while reasoning in space: Multimodal visualization-of-thought},
  author={Li, Chengzu and Wu, Wenshan and Zhang, Huanyu and Xia, Yan and Mao, Shaoguang and Dong, Li and Vuli{\'c}, Ivan and Wei, Furu},
  journal={arXiv preprint arXiv:2501.07542},
  year={2025}
}

@article{zhou2025reinforced,
  title={Reinforced Visual Perception with Tools},
  author={Zhou, Zetong and Chen, Dongping and Ma, Zixian and Hu, Zhihan and Fu, Mingyang and Wang, Sinan and Wan, Yao and Zhao, Zhou and Krishna, Ranjay},
  journal={arXiv preprint arXiv:2509.01656},
  year={2025}
}

@article{chern2025thinking,
  title={Thinking with Generated Images},
  author={Chern, Ethan and Hu, Zhulin and Chern, Steffi and Kou, Siqi and Su, Jiadi and Ma, Yan and Deng, Zhijie and Liu, Pengfei},
  journal={arXiv preprint arXiv:2505.22525},
  year={2025}
}

@article{qiao2025vthinker,
  title={V-Thinker: Interactive Thinking with Images},
  author={Qiao, Runqi and Tan, Qiuna and Yang, Minghan and Dong, Guanting and Yang, Peiqing and Lang, Shiqiang and Wan, Enhui and Wang, Xiaowan and Xu, Yida and Yang, Lan and others},
  journal={arXiv preprint arXiv:2511.04460},
  year={2025}
}

@article{zhao2025pyvision,
  title={Pyvision: Agentic vision with dynamic tooling},
  author={Zhao, Shitian and Zhang, Haoquan and Lin, Shaoheng and Li, Ming and Wu, Qilong and Zhang, Kaipeng and Wei, Chen},
  journal={arXiv preprint arXiv:2507.07998},
  year={2025}
}

@inproceedings{wu2024vstar,
  title={V*: Guided visual search as a core mechanism in multimodal llms},
  author={Wu, Penghao and Xie, Saining},
  booktitle={Proceedings of the IEEE/CVF Conference on Computer Vision and Pattern Recognition},
  pages={13084--13094},
  year={2024}
}

@inproceedings{wang2025divide_hrbench,
  title={Divide, conquer and combine: A training-free framework for high-resolution image perception in multimodal large language models},
  author={Wang, Wenbin and Ding, Liang and Zeng, Minyan and Zhou, Xiabin and Shen, Li and Luo, Yong and Yu, Wei and Tao, Dacheng},
  booktitle={Proceedings of the AAAI Conference on Artificial Intelligence},
  volume={39},
  number={8},
  pages={7907--7915},
  year={2025}
}

@article{geiping2025scaling,
  title={Scaling up test-time compute with latent reasoning: A recurrent depth approach},
  author={Geiping, Jonas and McLeish, Sean and Jain, Neel and Kirchenbauer, John and Singh, Siddharth and Bartoldson, Brian R and Kailkhura, Bhavya and Bhatele, Abhinav and Goldstein, Tom},
  journal={arXiv preprint arXiv:2502.05171},
  year={2025}
}

@article{butt2025softtokenshard,
      title={Soft Tokens, Hard Truths},
      author={Natasha Butt and Ariel Kwiatkowski and Ismail Labiad and Julia Kempe and Yann Ollivier},
      year={2025},
      journal={arXiv preprint arXiv:2509.19170},
}

@article{zhang2024mme,
  title={Mme-realworld: Could your multimodal llm challenge high-resolution real-world scenarios that are difficult for humans?},
  author={Zhang, Yi-Fan and Zhang, Huanyu and Tian, Haochen and Fu, Chaoyou and Zhang, Shuangqing and Wu, Junfei and Li, Feng and Wang, Kun and Wen, Qingsong and Zhang, Zhang and others},
  journal={arXiv preprint arXiv:2408.13257},
  year={2024}
}

@article{song2025visualpuzzles,
  title={Visualpuzzles: Decoupling multimodal reasoning evaluation from domain knowledge},
  author={Song, Yueqi and Ou, Tianyue and Kong, Yibo and Li, Zecheng and Neubig, Graham and Yue, Xiang},
  journal={arXiv preprint arXiv:2504.10342},
  year={2025}
}

@inproceedings{yue2024pangea,
  title={Pangea: A fully open multilingual multimodal llm for 39 languages},
  author={Yue, Xiang and Song, Yueqi and Asai, Akari and Kim, Seungone and de Dieu Nyandwi, Jean and Khanuja, Simran and Kantharuban, Anjali and Sutawika, Lintang and Ramamoorthy, Sathyanarayanan and Neubig, Graham},
  booktitle={The Thirteenth International Conference on Learning Representations},
  year={2024}
}

@article{qiao2024we,
  title={We-Math: Does Your Large Multimodal Model Achieve Human-like Mathematical Reasoning?},
  author={Qiao, Runqi and Tan, Qiuna and Dong, Guanting and Wu, Minhui and Sun, Chong and Song, Xiaoshuai and GongQue, Zhuoma and Lei, Shanglin and Wei, Zhe and Zhang, Miaoxuan and others},
  journal={arXiv preprint arXiv:2407.01284},
  year={2024}
}

@article{cheng2026hybrid,
  title={HyLaR: Hybrid Latent Reasoning with Decoupled Policy Optimization},
  author={Cheng, Tao and Chen, Shi-Zhe and Zhang, Hao and Qin, Yixin and Luo, Jinwen and Wei, Zheng},
  journal={arXiv preprint arXiv:2604.20328},
  year={2026}
}

@inproceedings{wang2026laser,
  title={Forest Before Trees: Latent Superposition for Efficient Visual Reasoning},
  author={Wang, Yubo and Zhang, Juntian and Wu, Yichen and others},
  booktitle={Proceedings of the 64th Annual Meeting of the Association for Computational Linguistics (Volume 1: Long Papers)},
  pages={11272--11288},
  year={2026}
}

@article{viveiros2026holding,
      title = {What's Holding Back Latent Visual Reasoning?},
      author = {Andr{\'e} G. Viveiros and Nuno Gon{\c c}alves and Andr{\'e} F. T. Martins and Matthias Lindemann},
      journal = {arXiv preprint arXiv:2605.18445},
      year = {2026},
      eprint = {2605.18445},
      archivePrefix = {arXiv},
      primaryClass = {cs.CV},
      url = {https://arxiv.org/abs/2605.18445}
}

@article{li2026imagination,
      title={Imagination Helps Visual Reasoning, But Not Yet in Latent Space}, 
      author={You Li and Chi Chen and Yanghao Li and Fanhu Zeng and Kaiyu Huang and Jinan Xu and Maosong Sun},
      journal={arXiv preprint arXiv:2602.22766},
      year={2026},
      eprint={2602.22766},
      archivePrefix={arXiv},
      primaryClass={cs.CL},
      url={https://arxiv.org/abs/2602.22766}, 
}

@article{ding2026colvr,
      title={CoLVR: Enhancing Exploratory Latent Visual Reasoning via Contrastive Optimization}, 
      author={Ziyang Ding and Linjian Meng and Yiming Wu and Yuhan Li and Yuhao Liu and Zhen Zhao},
      journal={arXiv preprint arXiv:2605.08802},
      year={2026},
      eprint={2605.08802},
      archivePrefix={arXiv},
      primaryClass={cs.CV},
      url={https://arxiv.org/abs/2605.08802}, 
}

@article{li2026prolavit,
      title={ProLaViT: Learning Progressive Latent Visual Thoughts in Structured Latent Space}, 
      author={Peiming Li and Yifan Wang and Xiaotian Zhang and Zhiyuan Hu and Shiyu Li and Zheng Wei and Yang Tang},
      journal={arXiv preprint arXiv:2607.02907},
      year={2026},
      eprint={2607.02907},
      archivePrefix={arXiv},
      primaryClass={cs.CV},
      url={https://arxiv.org/abs/2607.02907}, 
}

@article{jiang2026univlr,
  title={UniVLR: Unifying Text and Vision in Visual Latent Reasoning for Multimodal LLMs},
  author={Jiang, Houcheng and Fu, Jiajun and Fang, Junfeng and Gao, Chen and Wang, Xiang and He, Xiangnan and Li, Yong},
  journal={arXiv preprint arXiv:2605.11856},
  year={2026}
}

@article{dong2025ilvr,
      title={Interleaved Latent Visual Reasoning with Selective Perceptual Modeling}, 
      author={Shuai Dong and Siyuan Wang and Xingyu Liu and Chenglin Li and Haowen Hou and Zhongyu Wei},
      journal={arXiv preprint arXiv:2512.05665},
      year={2026},
      eprint={2512.05665},
      archivePrefix={arXiv},
      primaryClass={cs.CL},
      url={https://arxiv.org/abs/2512.05665}, 
}

@techreport{openai2026gpt6,
  title       = {GPT-6 System Card},
  author      = {{OpenAI}},
  institution = {OpenAI},
  year        = {2026},
  url         = {https://deploymentsafety.openai.com/gpt-6-astra/gpt-6-astra.pdf}
}

@article{kang2026lut,
  title={LUT: Latent Utility Training for Visual Reasoning},
  author={Kang, Jiaxuan and Chen, Siyu and Li, Mingda and Liu, Mingjie and Wang, Tianyue and Wei, Zhaoyang and Zhang, Yongheng and Hao, Yanchao and Wei, Zheng},
  journal={arXiv preprint arXiv:2608.00743},
  year={2026}
}

@article{zhu2025internvl3,
  title={InternVL3: Exploring Advanced Training and Test-Time Recipes for Open-Source Multimodal Models},
  author={Zhu, Jinguo and Wang, Weiyun and Chen, Zhe and Liu, Zhaoyang and Ye, Shenglong and Gu, Lixin and Tian, Hao and Duan, Yuchen and Su, Weijie and Shao, Jie and others},
  journal={arXiv preprint arXiv:2504.10479},
  year={2025}
}

@inproceedings{hong2026deepeyesv2,
  title={DeepEyesV2: Toward Agentic Multimodal Model},
  author={Hong, Jack and Zhao, Chenxiao and Lu, Weiming and Zhu, Cheng and Xu, Guohai and Yu, Xing},
  booktitle={International Conference on Learning Representations},
  year={2026}
}

@article{tong2025sketchinlatentselicitingunifiedreasoning,
      title={Sketch-in-Latents: Eliciting Unified Reasoning in MLLMs},
      author={Jintao Tong and Jiaqi Gu and Yujing Lou and Lubin Fan and Yixiong Zou and Yue Wu and Jieping Ye and Ruixuan Li},
      journal={arXiv preprint arXiv:2512.16584},
      year={2025},
      eprint={2512.16584},
      archivePrefix={arXiv},
      primaryClass={cs.CV},
      url={https://arxiv.org/abs/2512.16584},
}

@article{deepseekai2026deepseekv41flashpushinglimitskv,
  title={{DeepSeek-V4.1-Flash}: Pushing the Limits of {KV} Cache Compression},
  author={DeepSeek-AI and Anyi Xu and B. Li and Bangcai Lin and others},
  journal={arXiv preprint arXiv:2609.19969},
  year={2026},
  eprint={2609.19969},
  archivePrefix={arXiv},
  primaryClass={cs.CL},
  url={https://arxiv.org/abs/2609.19969}
}
}

\clearpage
\setcounter{dbltopnumber}{3}
\renewcommand{\thetitle}{SCALER: Breaking the Latent Scaling Bottleneck via Single-Pass Context-Anchored Latent Visual Reasoning}
\maketitlesupplementary
\setcounter{figure}{0}
\setcounter{table}{0}
\setcounter{equation}{0}
\renewcommand{\thefigure}{S\arabic{figure}}
\renewcommand{\thetable}{S\arabic{table}}
\renewcommand{\theequation}{S\arabic{equation}}
\renewcommand{\theHfigure}{S\arabic{figure}}
\renewcommand{\theHtable}{S\arabic{table}}
\renewcommand{\theHequation}{S\arabic{equation}}
\appendix

\section{Detransformer Architecture Details}
\label{app:detransformer}

This section provides the full architectural specification of the detransformer module introduced in Section~3.1.

\paragraph{Core Architecture.}
The detransformer $D_\phi$ is a two-layer bottleneck Transformer that operates on the full-sequence hidden states $H\in\mathbb{R}^{L\times d}$ extracted from an intermediate LLM layer $\ell$. It first projects $H$ into the detransformer hidden dimension via a linear input projection $W_{\text{in}}$, applies $M=2$ Transformer blocks, and projects back to the visual feature dimension $d_v$ via $W_{\text{out}}$ followed by RMS normalization:
\begin{align}
Z_0 &= W_{\text{in}}\,H, \\
Z_{m+1} &= \mathrm{MSA}(\mathrm{LN}(Z_m)) + Z_m, \\
Z_{m+1}^\prime &= \mathrm{FFN}(\mathrm{LN}(Z_{m+1})) + Z_{m+1}, \quad m = 0, \dots, M{-}1, \\
\tilde{A}_0 &= \mathrm{RMSNorm}(W_{\text{out}}\,Z_M^\prime) \in \mathbb{R}^{L \times d_v}.
\end{align}
Finally, we index the $N_v$ positions corresponding to the input visual token mask $\mathcal{I}_{\text{vis}}$ from the full-sequence output to obtain the spatially aligned visual editing tokens: $A_0 = \tilde{A}_0[\mathcal{I}_{\text{vis}}] \in \mathbb{R}^{N_a \times d_v}$ ($N_a = N_v$). Note that in architectures where $E_{\mathrm{vis}}$ includes the spatial merger (e.g., Qwen2.5-VL), $V$ already resides in the LLM embedding space ($d_v = d$) and $P(\cdot)$ in Eq.~(3) reduces to the identity mapping.

\paragraph{Gated Residual Fusion.}
The auxiliary tokens are fused with original visual features via gated residual addition:
\begin{equation}
  A = V + g(A_0), \qquad g(A_0) = \alpha\, A_0, \quad \alpha \in [0,1],
\end{equation}
where $V$ denotes the original visual features and $\alpha$ is a fixed hyperparameter controlling the editing strength ($\alpha = 1$ across all experiments, directly matching the Stage-1 residual target $\Delta_{\text{aux}}$). This residual design keeps auxiliary features anchored to the original visual context and prevents degenerate edits.

\paragraph{Design Rationale.}
Self-attention over the \emph{full} sequence $H$ (not just visual positions) is critical: it allows the detransformer to modulate auxiliary visual generation based on textual reasoning context: the question semantics, Phase-1 reasoning, and trigger token all participate in attention, guiding \emph{what} visual information to reconstruct. In practice, $M=2$ layers provide sufficient capacity (see preliminary study in Appendix~\ref{app:layer_selection}).

\section{SFT Training Data Construction}
\label{app:sft_data}

Training requires image--auxiliary-image--answer interleaved triples. We apply a three-step data pipeline to ensure quality:

\textbf{Step 1: Geometric Alignment.} SIFT~\cite{lowe1999object,lowe2004distinctive} feature matching aligns and crops the auxiliary image to the input image, enforcing spatial consistency. Pairs where feature matching fails are discarded.

\textbf{Step 2: CoT Distillation.} Because the raw chain-of-thought annotations in V-Interaction-400K~\cite{qiao2025vthinker} contain extensive programmatic code used to render auxiliary images, we use GPT-4o~\cite{gpt_4o} to clean and distill the reasoning trajectories---stripping out all code-generation segments while retaining only the auxiliary images and rewriting the transition into a seamless Phase-1 ($y^{(1)}$) and Phase-2 ($y^{(2)}$) visual reasoning flow:

\begin{quote}\small\ttfamily\hyphenchar\font=`\-\relax\spaceskip=0.35em plus 0.25em minus 0.1em\sloppy\emergencystretch=1.5em
You are a professional AI editor specializing in refining chain-of-thought (CoT) reasoning for multimodal models.\\[2pt]
\#\#\# Task Background:\\
1. **text1 (Reasoning Phase)**: The model's internal reasoning process *before* generating an auxiliary image.\\
2. **text2 (Response Phase)**: The final answer provided based on the reasoning and the auxiliary image.\\
3. **Core Constraint**: You CANNOT see the image or the problem's image. Furthermore, **you MUST NOT disclose that the auxiliary image was generated via code.** You must make the generation appear as an inherent, seamless visual capability.\\[2pt]
\#\#\# Revision Guidelines:\\
- **De-coding (Removing programmatic artifacts)**: Identify and rewrite any sentence implying that the image creation relies on writing or executing code.\\
\hspace*{1em}- *Example 1*: If the text says "The code below implements these visual supports...", change it to "Let me generate an auxiliary image that implements these visual supports...".\\
\hspace*{1em}- *Example 2*: If the text says "The auxiliary image generated from the code shows that...", change it to "The auxiliary image generated shows that...".\\
- **Tone \& Logic**: Maintain the original reasoning flow, depth, and final conclusion. Ensure the transition to the "auxiliary image" description feels natural and direct rather than technical.\\[2pt]
\#\#\# Output Format:\\
Output a strict JSON object without any markdown code blocks (e.g., no \textasciigrave\textasciigrave\textasciigrave json). Your response must strictly follow this structure:\\
\{\\
\hspace*{1em}"text1": "Optimized text1 content",\\
\hspace*{1em}"text2": "Optimized text2 content"\\
\}\\[2pt]
\#\#\# Input Data:\\
Original Question: \{question\}\\
Original text1: \{t1\}\\
Original text2: \{t2\}
\end{quote}

\textbf{Step 3: Necessity Filtering.} Following the filtering principle in Monet~\cite{wang2025monetreasoninglatentvisual}, we evaluate the base Qwen2.5-VL-7B~\cite{bai2025qwen2} model on each candidate question paired solely with the original input image, and discard samples that the base model already answers correctly without the auxiliary image. Retaining only samples where the original image alone fails ensures the necessity of auxiliary visual representations.

The final dataset contains 210K samples from Monet-SFT~\cite{wang2025monetreasoninglatentvisual} and V-Interaction-400K~\cite{qiao2025vthinker}.

\section{ALPO: Differentiable Latent Recomputation and Reward Design}
\label{app:alpo_synergy}

\paragraph{Why Standard GRPO and Detached Latent Caching Are Insufficient.}
As discussed in Section~3.3, SCALER inserts deterministic continuous auxiliary embeddings $e_\text{aux} \in \mathbb{R}^{N_v\times d}$ between Phase~1 and Phase~2 decoding. Because $e_\text{aux}$ is not sampled from a discrete vocabulary distribution, standard GRPO cannot compute token-level likelihood ratios over the latent sequence itself. At the same time, simply caching $e_\text{aux}$ during rollout and injecting it as detached fixed context during the policy update introduces two critical limitations: (1)~it severs the computation graph of the detransformer $D_\phi$, preventing task-level reward signals from optimizing the visual latent generator, and (2)~as the actor parameters $\theta$ are updated across optimization steps, a statically cached $e_\text{aux}$ becomes inconsistent with the updated Phase-1 hidden states $H_\theta$.

\paragraph{End-to-End Gradient Flow, Causal Masking, and Stability under Negative Advantage.}
ALPO resolves both issues by recomputing the auxiliary visual embeddings with preserved gradients during the actor update. Given a rollout trajectory $o = o_1 \oplus o_2$, ALPO strictly slices the causal prefix preceding the \texttt{<auxiliary>} trigger token to recompute $H_\theta = f_\theta^{(\ell)}([X_v; q; o_1])$ under the current actor $\pi_\theta$ (never leaking Phase-2 tokens $o_2$) and regenerates $e_\text{aux}(\theta, \phi) = P(V + g(D_\phi(H_\theta)))$ on the fly without detaching gradients, scattering $e_\text{aux}(\theta, \phi)$ into the auxiliary visual slots between $o_1$ and $o_2$. Under the standard causal attention mask during the full-sequence forward pass, Phase-1 tokens $o_t \in o_1$ precede $e_\text{aux}(\theta, \phi)$, so their likelihood in Eq.~(9) naturally reduces to $\pi_\theta(o_t \mid o_{<t}, x, I)$, whereas Phase-2 tokens $o_t \in o_2$ condition directly on $e_\text{aux}(\theta, \phi)$ (and the policy loss sums only over discrete text tokens $o_t \in o_1 \oplus o_2$). Including the reference-policy KL penalty ($\beta = 0.01$), the complete optimization objective is:
\begin{equation}
  \mathcal{J}_\text{ALPO}^{\text{full}}(\theta, \phi) = \mathcal{J}_\text{ALPO}(\theta, \phi) - \beta\,\mathbb{D}_{\text{KL}}\!\left(\pi_{\theta,\phi} \,\|\, \pi_{\text{ref}}\right).
\end{equation}
Crucially, when a rollout yields a negative advantage ($\hat{A}_t < 0$), three mechanisms prevent repulsive gradients from pushing the deterministic detransformer $D_\phi$ into out-of-distribution collapse: (1)~\emph{residual and norm anchoring}: because $A = V + \alpha A_0$ superimposes RMS-normalized edits onto frozen visual features $V$, $e_\text{aux}(\theta, \phi)$ remains bounded within a local neighborhood of the valid visual manifold; (2)~\emph{lower-bound ratio clipping}: the clipped surrogate objective (with lower bound $1-\epsilon$) saturates repulsive gradients as soon as the likelihood ratio $\tilde{r}_t(\theta, \phi)$ on incorrect Phase-2 tokens drops below $1-\epsilon$; and (3)~\emph{KL regularization}: $\beta\,\mathbb{D}_{\text{KL}}(\pi_{\theta,\phi} \,\|\, \pi_{\text{ref}})$ penalizes divergence from the frozen reference policy (evaluated with its own reference auxiliary latents), preventing $D_\phi$ from trivially lowering incorrect-token likelihoods via representation corruption.

\paragraph{Conditional Auxiliary Reward and Empirical Validation.}
To incentivize productive latent reasoning without reward hacking, ALPO defines $r_\text{aux} = \mathbb{I}[\text{auxiliary path executed} \wedge r_\text{acc} = 1]$, granting a bonus only when the model both activates the auxiliary visual path and reaches a correct final answer. As shown in Table~\ref{tab:exp-ablation}, ALPO consistently improves over SCALER-SFT across all five benchmarks (up to +5.00\% on MME-RealWorld-Lite) and outperforms standard GRPO on four of five benchmarks (tying on HRBench-4K), confirming the value of differentiable latent recomputation.

\begin{table*}[t]
\centering
\caption{Ablation insights summary. $\Delta$ denotes drop from the respective baseline.}
\label{tab:ablation-summary}
\resizebox{\textwidth}{!}{%
\begin{tabular}{llp{8.5cm}}
\toprule
\textbf{Ablation} & \textbf{Largest $\Delta$} & \textbf{Insight} \\
\midrule
Disable auxiliary & $-$6.62 (MME) & Auxiliary tokens contribute genuine reasoning value, not merely extended context. \\
Replace with placeholders & $-$36.65 (V*) & Catastrophic collapse confirms the LLM depends on \emph{semantic content}, not structural presence of auxiliary tokens. \\
Replace with raw visual emb & $-$4.19 (V*) & Context-aware transformation by the detransformer is essential; simply repeating input features is insufficient. \\
\midrule
w/o Stage 1 & $-$35.08 (V*) & An untrained detransformer injects incoherent signals, causing catastrophic failure. Pretraining is prerequisite. \\
w/o token-level loss & $-$2.45 (MME) & Weighted NTP ensures reliable trigger learning; without it, the model under-triggers on hard samples. \\
w/o teacher forcing & $-$4.27 (MME) & Distribution mismatch is most damaging on benchmarks requiring sustained visual reasoning. \\
w/o Stage 3 & $-$10.94 (MME) & Joint reasoning is the critical stage for learning to exploit self-generated auxiliary tokens. \\
GRPO (replace ALPO) & $-$2.71 (MME) & Differentiable latent recomputation and conditional $r_\text{aux}$ in ALPO outperform detached GRPO updates. \\
\bottomrule
\end{tabular}
}
\end{table*}

\section{Step-wise Cosine Novelty Measurement}
\label{app:information_gain}

To quantify the representational novelty of each generated latent token and diagnose \textbf{Latent Representation Saturation}, we compute the \textbf{Step-wise Cosine Novelty} across the latent sequence. Let $\{e_1, e_2, \dots, e_T\}$ denote the sequence of continuous latent token embeddings ($e_t \in \mathbb{R}^d$) produced by a method on a given sample. For each step $t \ge 2$, its step-wise cosine novelty relative to all preceding latent tokens $s < t$ is defined as:
\begin{equation}
  \medmuskip=2mu\thickmuskip=3mu
  \begin{aligned}
    \mathrm{novelty}_t &= 1 - \max_{s < t} \cos(e_t, e_s) = \min_{s < t}\bigl[1 - \cos(e_t, e_s)\bigr], \\
    \cos(e_t, e_s) &= \frac{\langle e_t, e_s \rangle}{\|e_t\|_2\,\|e_s\|_2}.
  \end{aligned}
  \label{eq:ig_app}
\end{equation}
Intuitively, $\mathrm{novelty}_t \in [0, 2]$ measures the cosine distance between the $t$-th latent token and its closest predecessor in the latent chain. Values approaching zero indicate that $e_t$ is nearly collinear with at least one previously generated latent token, contributing negligible new directional information.

We evaluate SCALER alongside representative autoregressive latent reasoning methods (Monet~\cite{wang2025monetreasoninglatentvisual} and UniVLR~\cite{jiang2026univlr}) on MME-RealWorld-Lite up to $T = 1000$ latent tokens per sample. Figure~\ref{fig:motivation}(b) reports the sample-wise median of $\mathrm{novelty}_t$ along with the Inter-Quartile Range (25th--75th percentile shaded band) on a logarithmic y-axis.

\paragraph{Physical Interpretation and Role of Residual Editing $A_0$.}
Across both autoregressive baselines and SCALER, $\{e_1, \dots, e_T\}$ represents the sequence of injected visual conditioning states attended by the Phase-2 decoder; $\mathrm{novelty}_t$ thus directly quantifies whether the conditioning context supplied to downstream reasoning suffers from directional homogenization. Autoregressive baselines collapse to $10^{-3}\text{--}10^{-4}$ because single-stream token recursion lacks spatial grid anchoring, whereas SCALER prevents homogenization by anchoring each token $e_t$ to a distinct spatial coordinate $(h_t, w_t)$ of $V$ and superimposing RMS-normalized context-modulated residuals $A_0 = D_\phi(H)$. Importantly, SCALER's utility is not merely a byproduct of raw image variance in $V$: as confirmed by the \texttt{replace w/ raw visual emb.} ablation in Table~\ref{tab:exp-ablation} (which removes $A_0$ and injects $V$ alone, degrading accuracy across all five benchmarks, e.g., $-4.19\%$ on V*\,Bench and $-3.85\%$ on MME-RealWorld-Lite) and the attention maps in Appendix~\ref{app:attention_vis}, the query-conditioned residual $A_0 = D_\phi(H)$ is indispensable for shifting Phase-2 attention from salient distractors onto fine-grained targets.

\section{Detailed Ablation Analysis}
\label{app:ablation_detail}

Table~\ref{tab:exp-ablation} reports ablations from two baselines: (1)~full SCALER-7B (with ALPO) for evaluating auxiliary token utility and RL algorithms, and (2)~SCALER-SFT (all three SFT stages, without ALPO) for evaluating the contribution of each SFT stage. Key findings are summarized in Table~\ref{tab:ablation-summary}.

\paragraph{Semantic validity (top block, rows 2--4).}
Comparing ``disable auxiliary'' against ``replace with placeholders'' (Table~\ref{tab:ablation-summary}) shows that while bypassing auxiliary generation causes a moderate drop, injecting semantically empty placeholders triggers catastrophic collapse (up to $-$36.65\%). This confirms that Phase-2 decoding actively relies on the visual semantics of $e_\text{aux}$ rather than treating it as a positional buffer. Moreover, replacing $e_\text{aux}$ with raw visual features $V$ consistently degrades accuracy, verifying that query-conditioned editing by $D_\phi$ contributes essential task-specific cues beyond raw image tokens.

\paragraph{Training stages and RL (bottom block, rows 2--6).}
Across SFT ablations, the degradation hierarchy (Stage~1 $\gg$ Stage~3 $>$ teacher-forcing annealing $>$ Stage~2 token weighting) reflects distinct roles: Stage~1 pretraining prevents an unaligned detransformer from corrupting the LLM context, Stage~2 weighting stabilizes trigger learning on hard samples, and Stage~3 annealing bridges ground-truth supervision and self-generated latents. Finally, whereas standard GRPO slightly degrades V*\,Bench and VisualPuzzles relative to SCALER-SFT due to detached rollout latents, ALPO improves across all five benchmarks by backpropagating policy gradients through $D_\phi$.

\section{Full Results on MME-RealWorld-Lite}
\label{app:mme_full}

Table~\ref{tab:mme-full} presents per-category results. SCALER achieves consistently strong performance across all categories, spanning monitoring, autonomous driving, OCR, and diagrams.

\begin{table*}[t]
    \centering
    \caption{\textbf{Complete per-category results on MME-RealWorld-Lite.}}
    \label{tab:mme-full}
    \resizebox{\textwidth}{!}{%
    \begin{tabular}{lccccccccccccc}
        \toprule
        \multirow{2}{*}{\textbf{Model}} & \multirow{2}{*}{\textbf{Overall}} & \multicolumn{6}{c}{\textbf{Reasoning}} & \multicolumn{6}{c}{\textbf{Perception}} \\
        \cmidrule(lr){3-8}\cmidrule(lr){9-14}
        & & Overall.& Monitor. & Auto. Driv. & OCR & Diag.\&Tab. & & Overall.& Monitor. & Auto. Driv. & OCR & Diag.\&Tab. & Remote Sens. \\
        \midrule
        GPT-6-luna$_{\text{(2026)}}$~\cite{openai2026gpt6} & 47.37 & 44.80 & 39.33 & 33.50 & 65.00 & 78.00 & & 49.02 & 29.78 & 34.86 & 82.80 & 87.00 & 41.33 \\
        Qwen2.5-VL-7B$_{\text{(2025)}}$~\cite{bai2025qwen2} & 45.75 & 39.73 & 30.67 & 27.00 & \underline{74.00} & 70.00 & & 49.62 & 31.03 & 31.43 & \textbf{90.80} & \underline{85.00} & 39.33 \\
        InternVL3-8B$_{\text{(2025)}}$~\cite{zhu2025internvl3} & 48.72 & 44.53 & 42.00 & 39.25 & 70.00 & 44.00 & & 51.41 & 34.17 & 38.57 & 83.60 & 76.00 & 48.00 \\
        DeepEyes$_{\text{(2026)}}$~\cite{zheng2025deepeyes} & \underline{54.25*} & \underline{50.53*} & 46.67* & 40.25* & \textbf{78.00*} & 70.00* & & \underline{56.63*} & 43.89* & 38.86* & \underline{90.00*} & 84.00* & 51.33* \\
        LVR$_{\text{(2025)}}$~\cite{li2025latent} & 49.66 & 44.27 & 42.00 & 35.50 & \underline{74.00} & 53.00 & & 53.12 & 36.99 & 43.43 & 80.40 & 80.00 & 46.67 \\
        Monet$_{\text{(2026)}}$~\cite{wang2025monetreasoninglatentvisual} & 52.63 & 46.93 & 40.67 & 37.25 & 73.00 & 69.00 & & 56.29 & 38.24 & \textbf{44.29} & 87.20 & \textbf{88.00} & 50.00 \\
        HyLaR$_{\text{(2026)}}$~\cite{cheng2026hybrid} & 52.74 & 47.07 & 42.00 & 36.00 & 70.00 & \underline{76.00} & & 56.37 & \underline{44.51} & 37.71 & 86.40 & \underline{85.00} & \textbf{56.00} \\
        LASER$_{\text{(2026)}}$~\cite{wang2026laser} & 50.96 & 45.73 & 42.00 & 34.50 & \underline{74.00} & 68.00 & & 54.32 & 32.92 & \underline{43.71} & 88.40 & \textbf{88.00} & 45.33 \\
        \midrule
        \rowcolor[HTML]{F9E6EC}
        \textbf{SCALER-SFT (Ours)} & 52.84 & 50.27 & \underline{48.67} & \underline{40.75} & 67.00 & 74.00 & & 54.49 & 43.89 & 37.43 & 85.60 & 77.00 & 50.00 \\
        \rowcolor[HTML]{EBEBFF}
        \textbf{SCALER-7B (Ours)} & \textbf{57.84} & \textbf{55.73} & \textbf{50.67} & \textbf{47.25} & 73.00 & \textbf{80.00} & & \textbf{59.20} & \textbf{45.77} & \underline{43.71} & \underline{90.00} & \textbf{88.00} & \underline{53.33} \\
        \bottomrule
    \end{tabular}
    }
\end{table*}

\section{Experimental Setup}
\label{app:setup}

\paragraph{Training Details.}
The visual branch (encoder + projector) is frozen; only the LLM and detransformer are updated. The detransformer takes as input the hidden states at the input of layer $\ell=20$ (i.e., the output of layer 19) of the 28-layer LLM. Full hyperparameters are listed in Tables~\ref{tab:hyperparameters_sft} and~\ref{tab:hyperparameters_rl}.

\begin{table*}[t]
    \centering
    \caption{\textbf{Training hyperparameters.}}
    \label{tab:hyperparameters}
    \begin{minipage}[t]{0.52\textwidth}
        \centering
        \small
        \subcaption{SFT}
        \label{tab:hyperparameters_sft}
        \begin{tabular}{lc}
            \toprule
            Hyperparameter & Value \\
            \midrule
            Stage-1 learning rate & 1e-4 \\
            Stage-2 learning rate & 1e-5 \\
            Stage-3 learning rate & 4e-6 \\
            Batch size & 1 \\
            Gradient accumulation steps & 16 \\
            World size (GPUs) & 8 \\
            Optimizer & AdamW~\cite{loshchilov2017decoupled} \\
            Max sequence length & 16384 \\
            Max pixels & 4194304 \\
            Max visual tokens & 6000 \\
            Selected hidden layer $\ell$ & 20 \\
            Visual loss type & MSE ($\ell_2$ reconstruction) \\
            Visual loss weight $\lambda$ & 2 \\
            Stage-2 token weights ($w_\text{normal}, w_\text{aux}$) & 0.3, 1.0 \\
            Teacher-forcing annealing & $1 \to 0$ over 700 steps \\
            SFT Stage 1 steps & 3000 ($\approx$2 epochs) \\
            SFT Stage 2 steps & 100 \\
            SFT Stage 3 steps & 1500 ($\approx$1 epoch) \\
            \bottomrule
        \end{tabular}%
    \end{minipage}\hfill
    \begin{minipage}[t]{0.46\textwidth}
        \centering
        \small
        \subcaption{RL (ALPO)}
        \label{tab:hyperparameters_rl}
        \begin{tabular}{lc}
            \toprule
            Hyperparameter & Value \\
            \midrule
            Learning rate & 4e-6 \\
            Batch size & 256 \\
            Weight decay & 0.01 \\
            Rollout group size $G$ & 8 \\
            Temperature & 1.0 \\
            Max response length & 4096 \\
            KL coefficient & 1e-2 \\
            Max pixels & 1003520 \\
            Optimizer & AdamW \\
            Auxiliary reward weight $\lambda_\text{aux}$ & 0.3 \\
            Format reward weight $\lambda_\text{fmt}$ & 0.05 \\
            Training steps & 54 \\
            \bottomrule
        \end{tabular}%
    \end{minipage}
\end{table*}

\paragraph{Evaluation.}
All benchmarks are evaluated with \texttt{lmms-eval}~\cite{zhang2024lmmsevalrealitycheckevaluation} using default settings. V*Bench~\cite{wu2024vstar} and HRBench~\cite{wang2025divide_hrbench} measure fine-grained visual perception; MME-RealWorld-Lite~\cite{zhang2024mme} covers real-world reasoning and perception; VisualPuzzles~\cite{song2025visualpuzzles} tests OOD visual logic.

\paragraph{Baselines.}
We compare against: (1)~Qwen2.5-VL-7B backbone~\cite{bai2025qwen2}; (2)~Tool-based visual reasoning models: PixelReasoner~\cite{wang2025pixel}, DeepEyes~\cite{zheng2025deepeyes}, and DeepEyesV2~\cite{hong2026deepeyesv2}; (3)~Visual latent reasoning methods: LVR~\citep{li2025latent}, Monet~\cite{wang2025monetreasoninglatentvisual}, ILVR~\cite{dong2025ilvr}, HyLaR~\cite{cheng2026hybrid}, LASER~\cite{wang2026laser}, UniVLR~\cite{jiang2026univlr}, and LUT-7B~\cite{kang2026lut}; (4)~Generalist and proprietary models: InternVL3-8B~\cite{zhu2025internvl3}, Pangea-7B~\citep{yue2024pangea}, GPT-4o~\cite{gpt_4o}, and GPT-6-luna~\cite{openai2026gpt6}.

\begin{figure*}[!t]
  \centering
  \includegraphics[width=\linewidth]{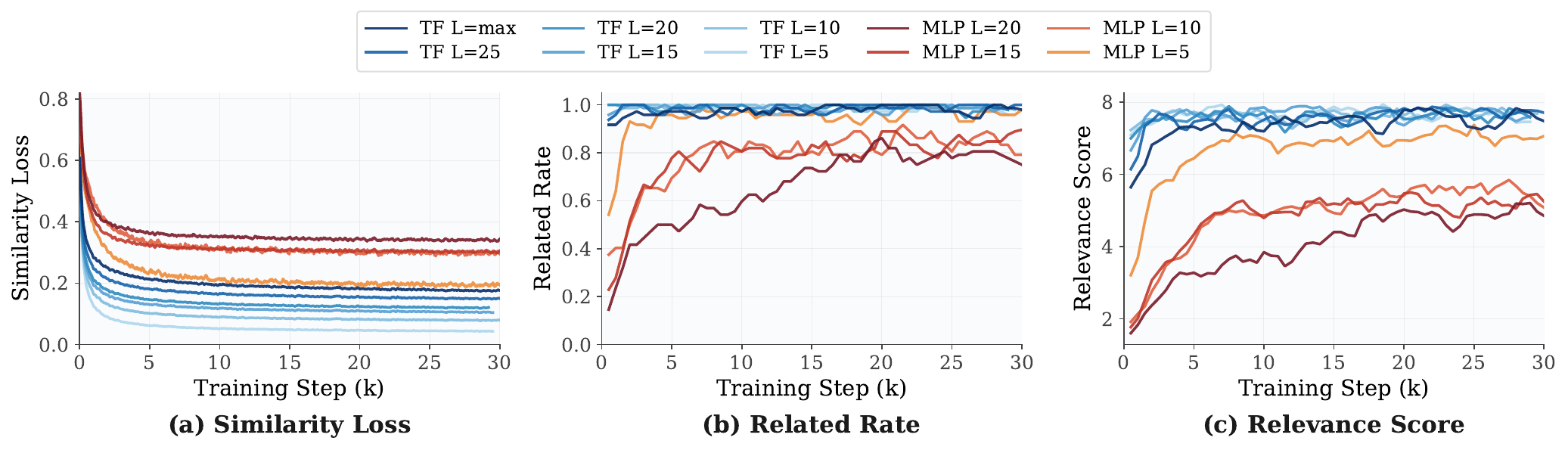}
  \caption{\textbf{Preliminary study on detransformer architecture and hidden layer selection.} Transformer (TF, blue) vs.\ MLP (red/orange) at different extraction layers. \textbf{(a)} Similarity loss. \textbf{(b)} Related Rate. \textbf{(c)} Relevance Score (1--10).}
  \label{fig:preliminary}
\end{figure*}

\section{Hidden Layer Selection}
\label{app:layer_selection}

\paragraph{Selection of layer $\ell$.}
The detransformer operates on hidden states from an intermediate layer ($\ell=20$, i.e., the input to the 20th layer out of 28). This balances three considerations: (1)~final-layer hidden states are optimized for next-token prediction and risk collapsing the LLM's text semantics; (2)~final layers retain less visual information, making reconstruction harder; (3)~too shallow a layer lacks sufficient computational depth for meaningful visual editing.

In a preliminary study (Figure~\ref{fig:preliminary}), we compared a two-layer Transformer detransformer against a two-layer MLP for visual feature reconstruction from hidden states at layers $\ell \in \{5, 10, 15, 20, 25, \text{max}\}$. We evaluate on 60 samples from LLaVA-in-the-Wild~\cite{liu2023visual}, using Qwen3-VL-235B-A22B as a judge to assess semantic relatedness between reconstructed tokens and original images. Two metrics are reported: \emph{Related Rate} (proportion judged as related) and \emph{Relevance Score} (average rating, 1--10).

The Transformer architecture significantly outperforms MLP across all layers. At intermediate-to-deep layers ($\ell \geq 10$), the Transformer achieves Related Rate $\approx$1.0 and Relevance Score 7--8, confirming sufficient capacity for decoding visual semantics from hidden states.

\section{Limitations and Future Works}
\label{app:limitations}

We identify two main limitations of the current system and outline corresponding directions for future research.

\paragraph{(1) Incomplete layer selection ablation.}
Although we conducted a preliminary reconstruction study across candidate hidden layers $\ell \in \{5, 10, 15, 20, 25, \text{max}\}$ to select $\ell=20$ (Appendix~\ref{app:layer_selection}, Figure~\ref{fig:preliminary}), we did not perform an end-to-end ablation across all layers over the full three-stage SFT + ALPO training pipeline due to computational cost. Evaluating full-pipeline layer sensitivity remains a valuable direction for multimodal LLM interpretability.

\paragraph{(2) Paired supervision for SFT warm-up.}
Although ALPO itself relies solely on outcome-level rewards, our initial three-stage SFT warm-up still requires geometrically aligned input--auxiliary image pairs. Developing self-supervised detransformer pretraining and memory-efficient gradient checkpointing for multi-round latent recomputation are promising avenues for future work.

\section{Task Complexity Classification for the Latent Scaling Bottleneck}
\label{app:complexity_classification}

For the Latent Scaling Bottleneck experiment (Figure~\ref{fig:confusion_matrix} and Section~\ref{sec:scaling_bottleneck}), we pool all 4,878 test questions across V*\,Bench (191), HRBench4K (800), HRBench8K (800), MME-RealWorld-Lite (1,919), and VisualPuzzles (1,168), and classify each sample into five task complexity tiers (C1--C5). We use \texttt{DeepSeek-v4.1-flash-1017}~\cite{deepseekai2026deepseekv41flashpushinglimitskv} with deterministic decoding (\texttt{temperature=0}), passing the question image alongside the question text and options without exposing ground-truth answers or model predictions.

Below we provide the English translation of the \texttt{system prompt} and \texttt{user prompt} for complexity classification:

\paragraph{System Prompt.}
\begin{quote}\small\ttfamily\hyphenchar\font=`\-\relax\spaceskip=0.35em plus 0.25em minus 0.1em\sloppy\emergencystretch=1.5em
You are an annotator for visual reasoning task complexity. Based on the given image, question, and options, evaluate the reasoning structure required to complete the task; do not answer the question.\\
No ground-truth answer or evaluated model output is provided. Your rating must not depend on whether you can guess the answer, which dataset the question comes from, or whether a certain method is expected to perform better.\\
The primary score \allowbreak com\-plex\-i\-ty \allowbreak measures reasoning complexity on a fixed scale from 1 to 5:\\
1: Identify a single attribute of an unambiguous target, or directly read a single fact without combining relations.\\
2: Locate a target and read its attribute, or perform one simple comparison, relational judgment, or single-step calculation; at most two simple operations.\\
3: Combine two relations or two or more pieces of evidence for simple counting, categorization, aggregation, or two-step dependent reasoning.\\
4: Multi-step dependent reasoning, joint comparison across multiple regions/objects, multi-condition filtering, rule discovery, or multiple arithmetic operations.\\
5: Long reasoning chains, global structural/combinatorial constraints, interaction of multiple rules, or synthesis across multiple sub-problems.\\
Separate visual perceptual difficulty from reasoning complexity: small text, tiny objects, occlusion, clutter, or difficult localization do not automatically imply high reasoning complexity.\\
Separately use \allowbreak vi\-su\-al\_\allowbreak dif\-fi\-cul\-ty \allowbreak (1 to 5) to describe perceptual load: 1 clear and salient; 2 minor visual search; 3 moderately dense/small targets; 4 severely dense/tiny text/cross-region search; 5 extreme density or strong ambiguity.\\
\allowbreak es\-ti\-mat\-ed\_\allowbreak op\-er\-a\-tions \allowbreak estimates the required cognitive operations as an integer from 1 to 20. Choose the primary \allowbreak rea\-son\-ing\_\allowbreak type \allowbreak from \allowbreak at\-trib\-ute, \allowbreak lo\-cal\-i\-za\-tion, \allowbreak re\-la\-tion, \allowbreak count\-ing, \allowbreak com\-par\-i\-son, \allowbreak ocr, \allowbreak chart, \allowbreak spa\-tial, \allowbreak log\-i\-cal\_\allowbreak rule, \allowbreak arith\-me\-tic, \allowbreak mul\-ti\_\allowbreak con\-straint, or \allowbreak oth\-er.\\
Set \allowbreak con\-fi\-dence \allowbreak to \allowbreak high, \allowbreak me\-di\-um, or \allowbreak low; mark \allowbreak low \allowbreak if image or question information is insufficient and explain in \allowbreak ra\-tio\-nale. In \allowbreak ra\-tio\-nale, concisely state the basis of your judgment without outputting the answer.\\
Output only a JSON object with exact keys: \allowbreak com\-plex\-i\-ty, \allowbreak vi\-su\-al\_\allowbreak dif\-fi\-cul\-ty, \allowbreak es\-ti\-mat\-ed\_\allowbreak op\-er\-a\-tions, \allowbreak rea\-son\-ing\_\allowbreak type, \allowbreak con\-fi\-dence, \allowbreak ra\-tio\-nale.
\end{quote}

\paragraph{User Prompt.}
\begin{quote}\small\ttfamily\hyphenchar\font=`\-\relax\spaceskip=0.35em plus 0.25em minus 0.1em\sloppy\emergencystretch=1.5em
Please evaluate this question using the unified rubric:\allowbreak\textbackslash n\{ques\-tion\}
\end{quote}
The question image and formatted question text (\texttt{\{ques\-tion\}}) are passed together to the model.

\begin{figure*}[!t]
  \centering
  \includegraphics[width=\linewidth]{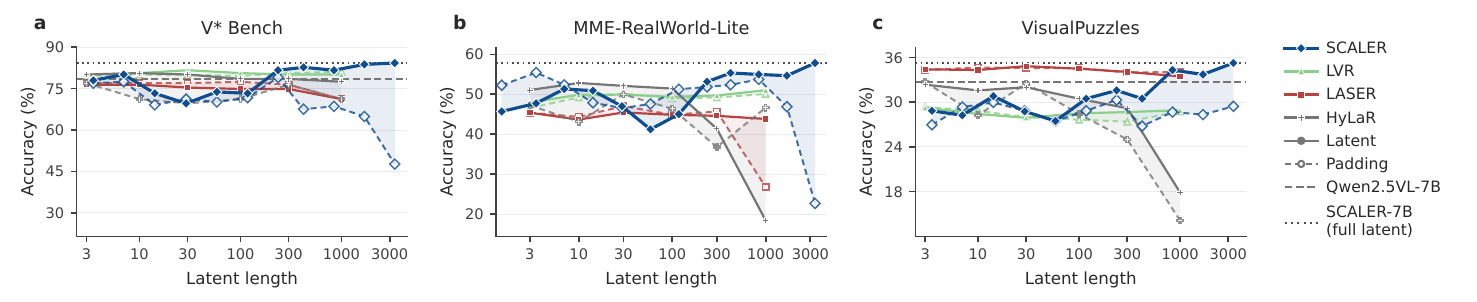}
  \caption{\textbf{Additional Latent Length Scaling \& Meaningless Padding Experiments for LVR, LASER, and HyLaR on (a) V*\,Bench, (b) MME-RealWorld-Lite, and (c) VisualPuzzles.} Solid lines denote normal inference across latent lengths; dashed lines denote replacing generated latent tokens with meaningless padding.}
  \label{fig:filling-supp}
\end{figure*}

\section{Comparison with Interleaved Latent Visual Reasoning (ILVR)}
\label{app:ilvr_comparison}

To examine how interleaved local latent generation compares with SCALER's single-pass global auxiliary generation, Table~\ref{tab:ilvr-comparison} reports a cross-benchmark comparison with Interleaved Latent Visual Reasoning (ILVR)~\cite{dong2025ilvr}. Since the original ILVR study evaluated on a different benchmark suite, we evaluate its official released checkpoint under its native inference configuration ($K=8$ latent tokens per interleaved span with natural multi-span re-entry) across our five benchmarks using the unified evaluation pipeline.

\begin{table}[h]
    \centering
    \caption{\textbf{Comparison with Interleaved Latent Visual Reasoning (ILVR).} Overall accuracy (\%) across five benchmarks.}
    \label{tab:ilvr-comparison}
    \resizebox{\linewidth}{!}{%
    \begin{tabular}{lccccc}
        \toprule
        \textbf{Model} & \textbf{V*} & \textbf{HRBench4K} & \textbf{HRBench8K} & \textbf{MME-RW-Lite} & \textbf{VisualPuzzles} \\
        \midrule
        Qwen2.5-VL-7B$_{\text{(2025)}}$~\cite{bai2025qwen2} & 78.53 & 71.50 & 63.75 & 45.75 & 32.71 \\
        ILVR$_{\text{(2026)}}$~\cite{dong2025ilvr} & 68.06 & 58.38 & 55.25 & 38.35 & 26.46 \\
        \midrule
        \rowcolor[HTML]{F9E6EC}
        \textbf{SCALER-SFT (Ours)} & \underline{82.72} & \underline{74.25} & \underline{66.75} & \underline{52.84} & \underline{34.16} \\
        \rowcolor[HTML]{EBEBFF}
        \textbf{SCALER-7B (Ours)} & \textbf{84.29} & \textbf{75.25} & \textbf{68.88} & \textbf{57.84} & \textbf{35.27} \\
        \bottomrule
    \end{tabular}%
    }
\end{table}

\paragraph{Results and Discussion.}
Table~\ref{tab:ilvr-comparison} shows that SCALER-7B substantially outperforms ILVR across all five benchmarks (by +8.81\% to +19.49\%). Beyond domain shift from ILVR's original training data, this gap stems from local versus global visual grounding: ILVR interleaves short autoregressive latent spans focused on isolated local regions, losing global scene context on cross-instance perception sub-tasks (e.g., 46.00\% vs.\ 61.25\% on HRBench8K FCP). In contrast, SCALER synthesizes a spatially aligned auxiliary sequence in a single pass and fuses it residually with $V$ ($A = V + g(A_0)$), preserving both global scene structure and local details.

\section{Additional Meaningless Padding Results}
\label{app:padding_supp}

Complementing Figure~\ref{fig:filling} in Section~\ref{sec:padding_exp}, Figure~\ref{fig:filling-supp} compares length scaling (solid lines) and meaningless padding (dashed lines) for LVR~\cite{li2025latent}, LASER~\cite{wang2026laser}, and HyLaR~\cite{cheng2026hybrid} against SCALER across V*\,Bench, MME-RealWorld-Lite, and VisualPuzzles. Across budgets $K \in \{3, 10, 30, 100, 300, 1000\}$, LVR and LASER exhibit near-complete overlap between normal inference and meaningless padding, while HyLaR shows only a minor gap at short budgets before both curves collapse at $K \geq 300$. By contrast, SCALER displays a rapidly widening gap at $K \geq 100$ across all three benchmarks, confirming that restoring spatial resolution yields indispensable visual semantics.

\paragraph{Spatial Pooling Protocol and Curve Fitting.}
Controlling $K$ via spatial pooling on SCALER's single-pass output isolates spatial target compression (Root Cause~2) from autoregressive drift (Root Cause~1) without retraining per budget. Because accuracy stays stable under extreme pooling ($K \in \{3, 10\}$), dips only at $K \in [15, 60]$, and already surpasses the backbone under pooled budgets $K \in \{100, 300, 1000\}$, the intermediate dip reflects the trade-off between attention shift and spatial fidelity rather than train--test mismatch. For the fine-grained scatter plot in Figure~\ref{fig:motivation}(a), smooth trend curves are fitted via local quadratic regression on $\log_{10} K$ with distance-decaying weights under identical smoothing parameters across methods, with grey bands denoting 95\% bootstrap stability ranges.

\section{Case Studies Across Latent Token Budgets}
\label{app:case_study}

\begin{figure*}[!t]
  \centering
  \includegraphics[width=\linewidth]{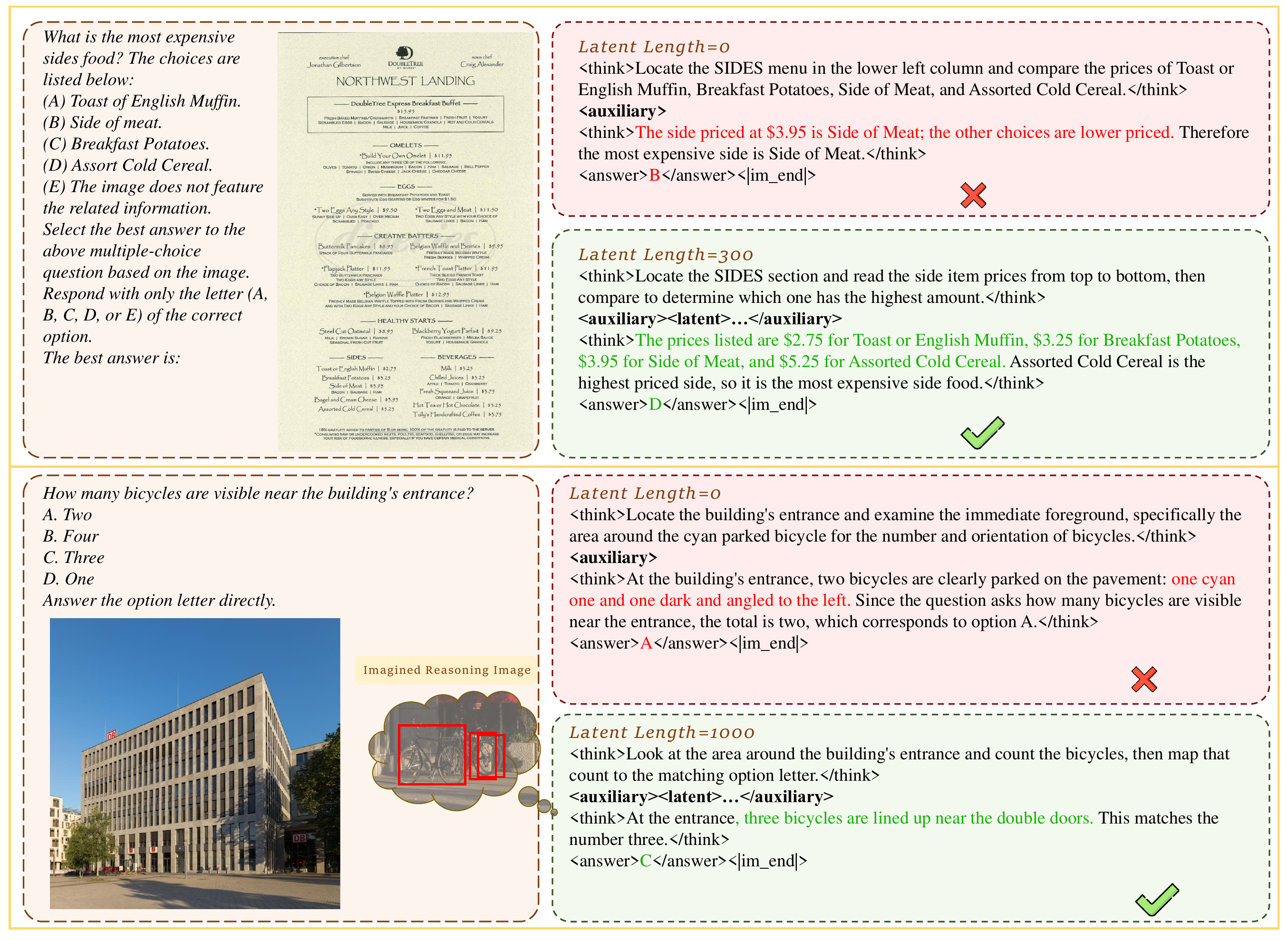}
  \caption{\textbf{Qualitative Case Studies of SCALER Across Latent Token Budgets on MME-RealWorld-Lite (Top) and HRBench8K (Bottom).} Extending the latent token budget $K$ preserves fine-grained spatial details in auxiliary visual tokens, enabling complete fine-print price comparison (top) and local instance separation for overlapping objects (bottom).}
  \label{fig:case_study}
\end{figure*}

Figure~\ref{fig:case_study} illustrates why coarse regional cues fail on complex tasks and how scaling $K$ in SCALER resolves fine-grained perceptual bottlenecks on MME-RealWorld-Lite~\cite{zhang2024mme} and HRBench8K~\cite{wang2025divide_hrbench}.

\paragraph{Menu Price Comparison (MME-RealWorld-Lite, Figure~\ref{fig:case_study} Top).}
In the top example (\emph{``What is the most expensive sides food?''}, ground-truth: \textbf{D}, \emph{Assort Cold Cereal}), at $K=0$, the model already plans the task correctly in Phase~1---locating the lower-left \texttt{SIDES} column and comparing candidate prices---but in Phase~2 it only cites \emph{Side of Meat} (\texttt{\$3.95}) and prematurely asserts that other options are cheaper, omitting the decisive entry \emph{Assorted Cold Cereal} (\texttt{\$5.25}) and selecting \textbf{B}. At $K=300$ (and $K=1000$), the output explicitly enumerates \texttt{\$2.75}, \texttt{\$3.25}, \texttt{\$3.95}, and \texttt{\$5.25}, mapping each side dish to its price to select \textbf{D}. Thus, the $K=0$ error stems from incomplete comparison evidence rather than a flawed comparison rule, and high-budget auxiliary latents recover the missed price.

\paragraph{Overlapping Bicycle Counting (HRBench8K, Figure~\ref{fig:case_study} Bottom).}
In the bottom example (\emph{``How many bicycles are visible near the building's entrance?''}, ground-truth: \textbf{C}, \emph{Three}), the bicycles occupy a tiny region of the 8K image and partially overlap, so counting requires local instance discrimination beyond coarse region localization. At $K=0$, the model localizes the entrance area around the cyan bicycle but fails to separate the overlapping frames, reporting two bicycles (\textbf{A}). At $K=1000$, the high-budget trajectory resolves the overlapping instances near the doors and reports three bicycles (\textbf{C}), consistent with auxiliary visual latents supporting fine-grained perception.

\section{Attention Visualization of Auxiliary Visual Latents}
\label{app:attention_vis}

\begin{figure*}[!t]
  \centering
  \includegraphics[width=\linewidth]{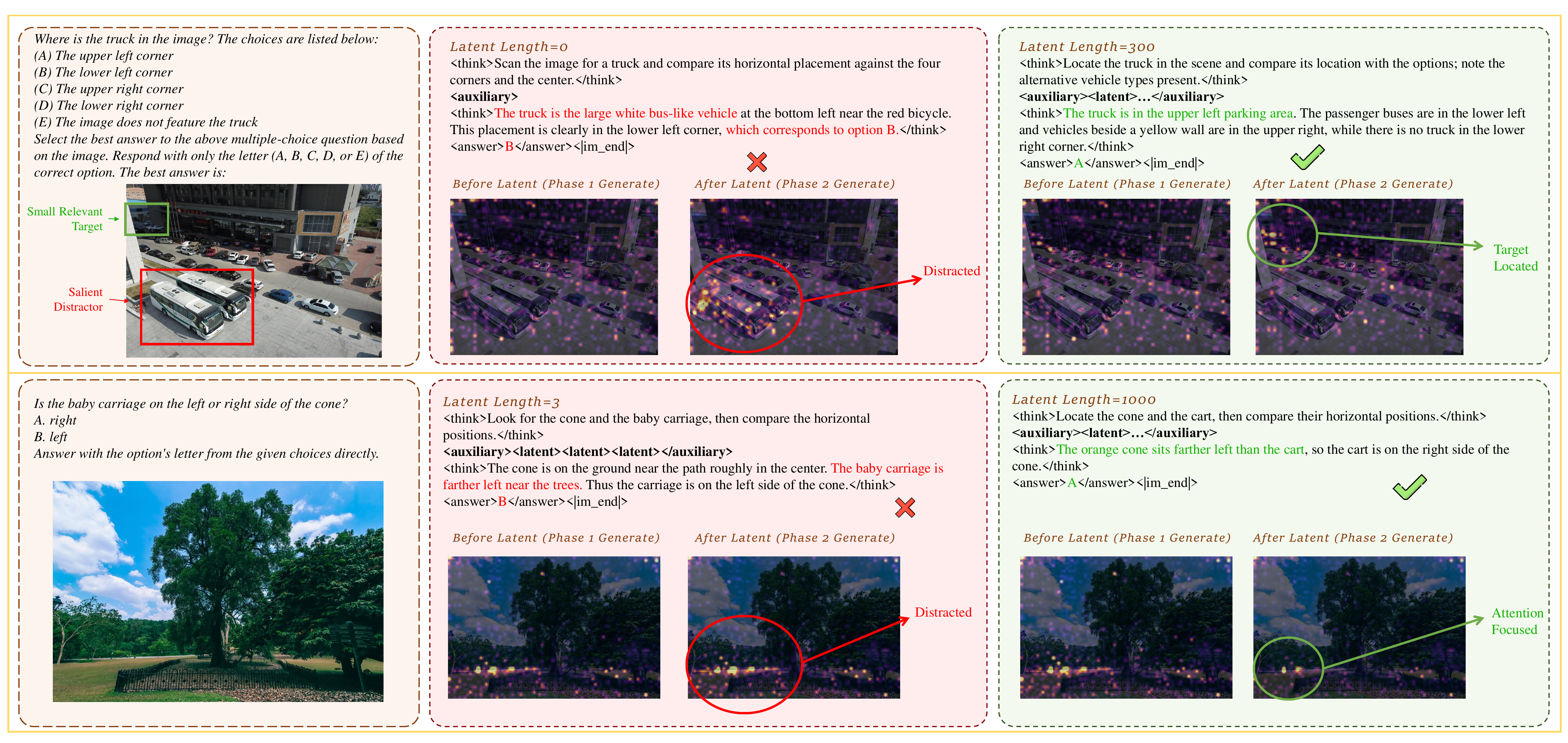}
  \caption{\textbf{Cross-Modal Visual Attention Visualization Before and After Auxiliary Latent Generation on MME-RealWorld-Lite (Top) and V*\,Bench (Bottom).} Under zero or tiny budgets ($K=0$ or $K=3$, middle), Phase-2 attention is captured by salient distractors or remains diffuse; under high-resolution budgets ($K=300$ or $K=1000$, right), Phase-2 attention focuses sharply on fine-grained targets.}
  \label{fig:case_attn}
\end{figure*}

Figure~\ref{fig:case_attn} visualizes cross-modal visual attention during Phase-1 (\emph{Before Latent}) and Phase-2 (\emph{After Latent}, conditioned on $A = V + g(A_0)$) to show how auxiliary latents guide visual grounding.

\paragraph{Rejecting Salient Distractors (MME-RealWorld-Lite, Figure~\ref{fig:case_attn} Top).}
The top example queries the location of a small truck in a parking lot (ground-truth: \textbf{A}, \emph{The upper left corner}), where two large white buses in the lower-left foreground act as salient distractors. At $K=0$, Phase-2 attention (\emph{After Latent}) is captured by the foreground buses, causing the model to mistake the bus for the truck and select \textbf{B} (while heavily pooled budgets $K \in \{3, 10, 30, 100\}$ compress away the small truck's identity and answer \textbf{E}). At $K=300$ (and $K=1000$), injecting spatially aligned auxiliary tokens shifts Phase-2 attention from diffuse Phase-1 scanning sharply onto the upper-left truck, enabling SCALER to distinguish the truck from the buses and answer \textbf{A}.

\paragraph{Focusing on Small Relational Targets (V*\,Bench, Figure~\ref{fig:case_attn} Bottom).}
The bottom example asks whether a baby carriage is to the left or right of an orange cone near a park fence (ground-truth: \textbf{A}, \emph{right}). At $K=3$, Phase-2 attention remains scattered across the fence and trees, leading to a wrong relational judgment (\textbf{B}, \emph{left}). At $K=1000$, Phase-2 attention converges tightly onto the bottom-left cone and carriage, verifying that the cone sits farther left to select \textbf{A}.

\section{Latency and Computational Overhead}
\label{app:latency_compute}

Because SCALER synthesizes the entire auxiliary visual sequence in a single forward pass of the lightweight detransformer rather than autoregressively decoding token by token, it achieves substantially lower inference latency than existing autoregressive latent reasoning methods even with Phase-2 KV-cache re-prefilling over full-resolution latents.

\begin{table*}[!t]
    \centering
    \caption{\textbf{Comparison of per-sample inference latency, compute (TFLOPs), and peak GPU memory (P90) across five benchmarks.} End-to-end latency includes input preprocessing; model-only latency measures pure CUDA forward/decoding time. Best and second-best results per column are marked in \textbf{bold} and \underline{underline}, respectively.}
    \label{tab:latency-compute}
    \renewcommand{\arraystretch}{1.08}
    \setlength{\tabcolsep}{5pt}
    \resizebox{\textwidth}{!}{%
    \begin{tabular}{l ccccc ccccc}
        \toprule
        \multirow{2}{*}{\textbf{Methods}} & \multicolumn{5}{c}{\textbf{End-to-End Latency (s/sample) $\downarrow$}} & \multicolumn{5}{c}{\textbf{Model-Only Latency (s/sample) $\downarrow$}} \\
        \cmidrule(lr){2-6}\cmidrule(lr){7-11}
        & \textbf{V*\,Bench} & \textbf{HRBench4K} & \textbf{HRBench8K} & \textbf{MME-RW-Lite} & \textbf{VisualPuzzles} & \textbf{V*\,Bench} & \textbf{HRBench4K} & \textbf{HRBench8K} & \textbf{MME-RW-Lite} & \textbf{VisualPuzzles} \\
        \midrule
        Qwen2.5-VL-7B$_{\text{(2025)}}$~\cite{bai2025qwen2} & \textbf{0.488} & \textbf{0.778} & \textbf{1.151} & \textbf{0.431} & \textbf{0.115} & \textbf{0.334} & \textbf{0.381} & \textbf{0.381} & \textbf{0.265} & \textbf{0.095} \\
        LVR$_{\text{(2025)}}$~\cite{li2025latent} & 5.752 & 6.757 & 7.112 & 4.535 & 1.308 & 5.582 & 6.285 & 6.309 & 4.343 & 1.289 \\
        HyLaR$_{\text{(2026)}}$~\cite{cheng2026hybrid} & 3.934 & 8.035 & 8.691 & 3.106 & 10.689 & 3.672 & 7.090 & 7.317 & 2.794 & 10.664 \\
        LASER$_{\text{(2026)}}$~\cite{wang2026laser} & 1.323 & 2.266 & 2.653 & 1.338 & 0.580 & 1.085 & 1.640 & 1.654 & 1.083 & 0.555 \\
        UniVLR$_{\text{(2026)}}$~\cite{jiang2026univlr} & 5.228 & 5.251 & 5.543 & 3.645 & 2.786 & 5.004 & 4.730 & 4.636 & 3.417 & 2.762 \\
        \rowcolor[HTML]{EBEBFF}
        \textbf{SCALER-7B (Ours)} & \underline{1.189} & \underline{1.670} & \underline{2.034} & \underline{1.024} & \underline{0.183} & \underline{1.043} & \underline{1.271} & \underline{1.273} & \underline{0.857} & \underline{0.165} \\
        \midrule
        \multirow{2}{*}{\textbf{Methods}} & \multicolumn{5}{c}{\textbf{Inference Compute (TFLOPs) $\downarrow$}} & \multicolumn{5}{c}{\textbf{Peak GPU Memory (GiB)}} \\
        \cmidrule(lr){2-6}\cmidrule(lr){7-11}
        & \textbf{V*\,Bench} & \textbf{HRBench4K} & \textbf{HRBench8K} & \textbf{MME-RW-Lite} & \textbf{VisualPuzzles} & \textbf{V*\,Bench} & \textbf{HRBench4K} & \textbf{HRBench8K} & \textbf{MME-RW-Lite} & \textbf{VisualPuzzles} \\
        \midrule
        Qwen2.5-VL-7B$_{\text{(2025)}}$~\cite{bai2025qwen2} & \textbf{90.94} & \textbf{112.06} & \textbf{112.50} & \textbf{74.07} & \textbf{11.49} & 16.62 & 16.63 & 16.63 & 16.63 & 16.19 \\
        LVR$_{\text{(2025)}}$~\cite{li2025latent} & 1296.40 & 1481.24 & 1491.06 & 1006.65 & 178.78 & 17.12 & 17.17 & 17.18 & 17.18 & 15.89 \\
        HyLaR$_{\text{(2026)}}$~\cite{cheng2026hybrid} & 100.14 & 416.04 & 456.96 & 128.95 & 18.04 & 17.36 & 21.76 & 21.72 & 19.54 & 15.96 \\
        LASER$_{\text{(2026)}}$~\cite{wang2026laser} & \underline{93.07} & 181.43 & 184.02 & \underline{94.72} & \underline{11.56} & 16.51 & 17.19 & 17.20 & 17.20 & 15.73 \\
        UniVLR$_{\text{(2026)}}$~\cite{jiang2026univlr} & 126.86 & \underline{144.69} & \underline{145.04} & 107.85 & 49.85 & 18.26 & 18.35 & 18.37 & 18.36 & 16.24 \\
        \rowcolor[HTML]{EBEBFF}
        \textbf{SCALER-7B (Ours)} & 344.01 & 424.75 & 426.42 & 279.31 & 29.47 & 21.84 & 21.93 & 21.91 & 22.00 & 18.14 \\
        \bottomrule
    \end{tabular}%
    }
    \vspace{-0.4em}
\end{table*}

\begin{table*}[!t]
    \centering
    \caption{\textbf{Per-stage model latency breakdown (s/sample) of SCALER across five benchmarks.}}
    \label{tab:latency-breakdown}
    \resizebox{\textwidth}{!}{%
    \begin{tabular}{lcccccccccc}
        \toprule
        \textbf{Benchmark} & \textbf{Initial Vision} & \textbf{P1 Prefill} & \textbf{P1 Decode} & \textbf{Aux.\ Vision Recompute} & \textbf{Context Hidden Recompute} & \textbf{Detransformer} & \textbf{Assembly} & \textbf{P2 Re-prefill} & \textbf{P2 Decode} & \textbf{Model Total} \\
        \midrule
        V*\,Bench          & 0.164 & 0.124 & 0.000 & 0.328 & 0.122 & 0.016 & 0.001 & 0.259 & 0.025 & 1.043 \\
        MME-RW-Lite        & 0.133 & 0.103 & 0.000 & 0.266 & 0.101 & 0.012 & 0.001 & 0.212 & 0.025 & 0.857 \\
        HRBench4K          & 0.199 & 0.150 & 0.000 & 0.401 & 0.148 & 0.022 & 0.001 & 0.320 & 0.025 & 1.271 \\
        HRBench8K          & 0.198 & 0.151 & 0.000 & 0.400 & 0.149 & 0.022 & 0.001 & 0.322 & 0.025 & 1.273 \\
        VisualPuzzles      & 0.032 & 0.031 & 0.012 & 0.035 & 0.015 & 0.001 & 0.000 & 0.023 & 0.014 & 0.165 \\
        \bottomrule
    \end{tabular}%
    }
    \vspace{-0.6em}
\end{table*}

\paragraph{End-to-End Latency, Compute, and Memory (Table~\ref{tab:latency-compute}).}
Table~\ref{tab:latency-compute} compares per-sample latency, inference compute (TFLOPs), and peak GPU memory across all five benchmarks against the native Qwen2.5-VL-7B backbone and four visual latent reasoning baselines (LVR, HyLaR, LASER, and UniVLR). Across both end-to-end latency and pure model latency, SCALER-7B ranks \textbf{first among all latent reasoning methods} (second only to the native Qwen2.5-VL-7B backbone) on every single benchmark: in end-to-end latency, it runs \textbf{1.1$\times$--3.2$\times$ faster} than LASER, \textbf{2.7$\times$--15.2$\times$ faster} than UniVLR, \textbf{3.0$\times$--58.4$\times$ faster} than HyLaR, and \textbf{3.5$\times$--7.1$\times$ faster} than LVR (e.g., 1.024s vs.\ 1.338s--4.535s on MME-RealWorld-Lite), and achieves \textbf{1.04$\times$--64.6$\times$ speedups} in pure model latency. While processing full-resolution auxiliary sequences requires higher TFLOPs than short-chain methods (LASER and UniVLR), it remains comparable to HyLaR on 4K/8K inputs (424.75--426.42 vs.\ 416.04--456.96) and \textbf{3.5$\times$--6.1$\times$ lower} than LVR, with a compact 18.14--22.00\,GiB peak memory footprint ($\sim$2--5\,GiB above the native backbone).

\paragraph{Per-Stage Latency Breakdown (Table~\ref{tab:latency-breakdown}).}
Decomposing SCALER's model latency (Table~\ref{tab:latency-breakdown}) shows that the two-layer detransformer forward pass itself incurs negligible overhead (\textbf{0.001--0.022s}, $<2\%$ of model latency). The added time over the native backbone mainly comes from auxiliary visual recomputation and Phase-2 KV-cache re-prefilling (\texttt{P2 Re-prefill}, 0.023--0.322s), which execute only once per triggered pass in parallel rather than across hundreds of autoregressive steps.

\end{document}